\documentclass{opendatalab}
\newcommand\blfootnote[1]{%
  \begingroup
  \renewcommand\thefootnote{}\footnote{#1}%
  \addtocounter{footnote}{-1}%
  \endgroup
}
\usepackage[utf8]{inputenc}
\usepackage{xcolor}
\usepackage[most]{tcolorbox}
\usepackage{longtable}
\usepackage{listings}
\usepackage[numbers]{natbib}
\usepackage{enumitem}
\usepackage{graphicx}
\usepackage{xspace} 
\usepackage{hyperref}

\usepackage{booktabs}        
\usepackage{graphicx}        
\usepackage[table]{xcolor}   
\usepackage{pifont}          
\usepackage{booktabs}   
\usepackage{multirow}   
\usepackage{graphicx}   
\usepackage{tabularx}

\usepackage{amsmath} 

\usepackage{titletoc}

\usepackage{booktabs, makecell, colortbl, xcolor}

\usepackage{algpseudocode}
\usepackage{algorithm}
\usepackage{xcolor}

\usepackage{tcolorbox}
\tcbuselibrary{skins,breakable}
\usepackage{wrapfig}

\usepackage{listings}
\usepackage{xcolor}

\usepackage{caption}

\usepackage[utf8]{inputenc} 
\usepackage[T1]{fontenc}    
\usepackage{hyperref}       
\usepackage{url}            
\usepackage{booktabs}       
\usepackage{amsfonts}       
\usepackage{nicefrac}       
\usepackage{microtype}      
\usepackage{xcolor}         
\definecolor{codegreen}{rgb}{0,0.6,0}
\definecolor{codegray}{rgb}{0.5,0.5,0.5}
\definecolor{codepurple}{rgb}{0.58,0,0.82}
\definecolor{backcolour}{rgb}{0.95,0.95,0.92}
\definecolor{promptcolor}{HTML}{D1D0F2}
\definecolor{promptcolorheader}{HTML}{bdbcec}

\newcommand{\github}{\raisebox{-1.5pt}{\includegraphics[height=1.05em]{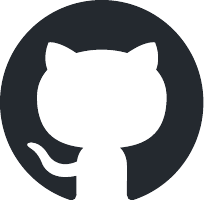}}\xspace}

\newcommand{\huggingface}{\raisebox{-1.5pt}{\includegraphics[height=1.05em]{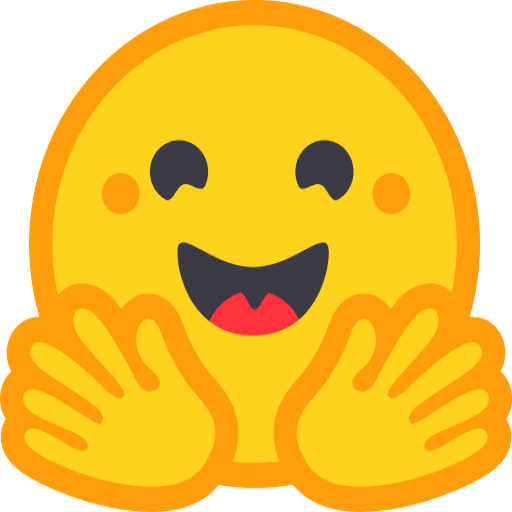}}\xspace}

\definecolor{promptcolor}{HTML}{E3F0FA}
\definecolor{promptcolorheader}{HTML}{B5D6ED}
\definecolor{prompttitletext}{HTML}{1B3A5C}

\newtcolorbox{promptbox}[1][]{
  enhanced, breakable,
  top=0.3em,bottom=0.3em,left=0.5em,right=0.5em,
  toptitle=0.3em,bottomtitle=0.2em,boxsep=0pt,
  colframe=promptcolorheader, colback=promptcolor!50, boxrule=0.5pt,
  width=\columnwidth, 
  coltitle=prompttitletext,
  title={\footnotesize #1} 
}
\lstdefinestyle{promptstyle}{
    backgroundcolor=\color{backcolour},   
    commentstyle=\color{codegreen},
    keywordstyle=\color{magenta},
    numberstyle=\tiny\color{codegray},
    stringstyle=\color{codepurple},
    basicstyle=\ttfamily\footnotesize,
    breakatwhitespace=false,         
    breaklines=true,                 
    captionpos=b,                    
    keepspaces=true,                 
    numbers=left,                    
    numbersep=5pt,                  
    showspaces=false,                
    showstringspaces=false,
    showtabs=false,                  
    tabsize=2
}
\title{NanoResearch: Co-Evolving Skills, Memory, and Policy for Personalized Research Automation}

\author[1]{Jinhang Xu$^\dagger$}
\author[1,2]{Qiyuan Zhu$^\dagger$}
\author[1,3]{Yujun Wu$^\dagger$}
\author[1,4]{Zirui Wang$^\dagger$}
\author[1,5]{Dongxu Zhang$^\dagger$}
\author[6]{Marcia Tian}
\author[1]{Yiling Duan}
\author[4]{Siyuan Li}
\author[1]{Jingxuan Wei}
\author[2]{Sirui Han$^{*}$}
\author[2]{Yike Guo$^{*}$}
\author[7]{Odin Zhang$^{*}$}
\author[1]{Conghui He$^{*}$}
\author[1]{Cheng Tan$^{*}$}

\affiliation[1]{Shanghai Artificial Intelligence Laboratory}
\affiliation[2]{The Hong Kong University of Science and Technology}
\affiliation[3]{Peking University}
\affiliation[4]{Zhejiang University}
\affiliation[5]{Xi'an Jiaotong University}
\affiliation[6]{East China University of Science and Technology}
\affiliation[7]{The Chinese University of Hong Kong}

\abstract{
LLM-powered multi-agent systems can now automate the full research pipeline from ideation to paper writing, but a fundamental question remains: automation for whom? Researchers operate under different resource configurations, hold different methodological preferences, and target different output formats. A system that produces uniform outputs regardless of these differences will systematically under-serve every individual user, making personalization a precondition for research automation to be genuinely usable. However, achieving it requires three capabilities that current systems lack: accumulating reusable procedural knowledge across projects, retaining user-specific experience across sessions, and internalizing implicit preferences that resist explicit formalization. We propose NanoResearch, a multi-agent framework that addresses these gaps through tri-level co-evolution. A skill bank distills recurring operations into compact procedural rules reusable across projects. A memory module maintains user- and project-specific experience that grounds planning decisions in each user's research history. A label-free policy learning converts free-form feedback into persistent parameter updates of the planner, reshaping subsequent coordination. These three layers co-evolve: reliable skills produce richer memory, richer memory informs better planning, and preference internalization continuously realigns the loop to each user. Extensive experiments demonstrate that NanoResearch delivers substantial gains over state-of-the-art AI research systems, and progressively refines itself to produce better research at lower cost over successive cycles.

\par\vspace{0.8em}
\noindent\makebox[\linewidth][c]{%
  \github~\href{https://github.com/OpenRaiser/NanoResearch}{\textbf{Code}}
  \qquad
  \huggingface~\href{https://huggingface.co/datasets/xjh111/nanoresearch-20topics}{\textbf{Dataset}}
}
\par\vspace*{-2.5em}
}

\begin{document}

\maketitle

\blfootnote{$^{\dagger}$Equal contribution.\quad $^{*}$Corresponding authors.}

\section{Introduction}

LLM-powered multi-agent systems~\cite{achiam2023gpt} have recently transformed end-to-end research automation from a long-standing aspiration~\cite{langley1987scientific, waltz2009automating} into working reality. Systems such as The AI Scientist~\cite{lu2024ai}, AI Scientist-v2~\cite{yamada2025ai}, EvoScientist~\cite{lyu2026evoscientist}, and AI-Researcher~\cite{tang2025ai} can now autonomously traverse the full research lifecycle~\cite{weng2025deepscientist, shao2025omniscientist, li2026autosota}, surveying literature, generating hypotheses, implementing experiments, and writing papers within a single pipeline. These advances mark genuine progress: tasks that once required weeks of researcher effort can now be completed in hours at modest cost~\cite{zhu2025ai}. Yet the ability to complete the pipeline does not guarantee that its outputs are usable by any particular researcher.

Research is fundamentally shaped by the context in which it is conducted~\cite{kuhn1970structure,latour2013laboratory}. Communities diverge in what constitutes a valuable contribution: AI-for-science researchers prioritize whether a method addresses a meaningful real-world need~\cite{moor2023foundation,wornow2023shaky}, while core computer vision researchers value architectural novelty and consistent benchmark gains~\cite{lipton2019troubling}. Beyond research philosophy, teams also differ in resource budgets~\cite{schwartz2020green}, methodological preferences, and target venues. A system that produces the same research plan regardless of these differences is unlikely to serve either community well. \textbf{Personalization is therefore a precondition for research automation to be genuinely usable}.

\begin{figure}[t]
  \includegraphics[width=\textwidth]{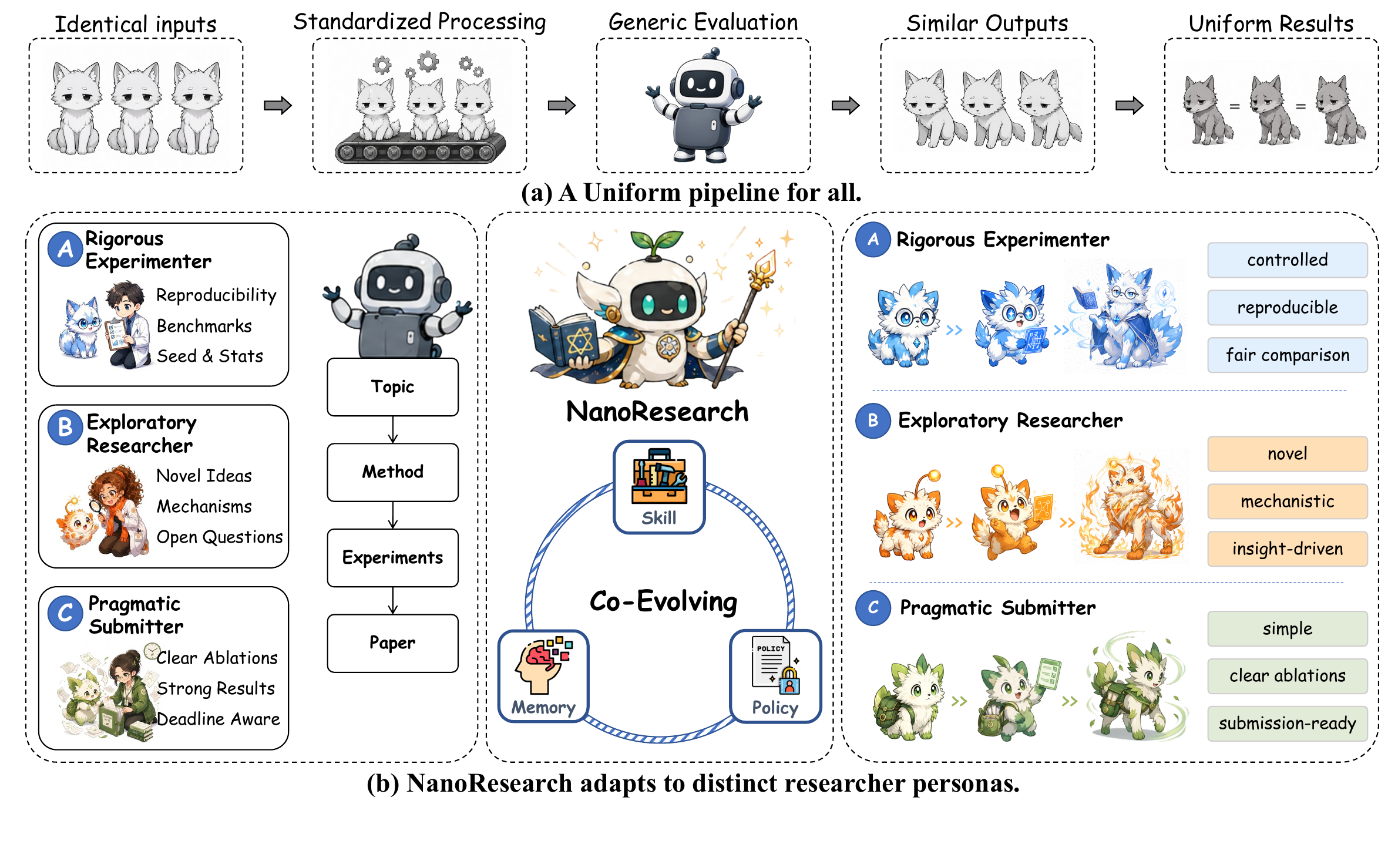} 
  \caption{Comparison between (a) a uniform research automation pipeline that applies identical processing to all users and yields homogeneous outputs, and (b) NanoResearch, which recognizes distinct researcher personas and provides personalized skills and feedback upon failure, enabling each persona to evolve along its own trajectory.}
\label{fig:teaser}
\end{figure}

Despite this need, existing systems remain fundamentally one-size-fits-all, funneling diverse researchers through a uniform pipeline that produces near-identical outputs regardless of individual context, as shown in Figure~\ref{fig:teaser}(a). We identify three capability gaps that jointly prevent personalization: \textit{(i) current systems lack reusable procedural knowledge}. Each run starts from scratch, re-encountering the same debugging patterns and re-deriving the same configurations without abstracting them into compact, retrievable rules. Even memory-equipped systems such as EvoScientist~\cite{lyu2026evoscientist} store episode-level narratives rather than distilled procedural primitives, limiting transferability across tasks. \textit{(ii) current systems do not accumulate user-specific experience across sessions}. Past hypotheses, validated configurations, and inferred resource constraints are discarded once a session ends, forcing rediscovery on every subsequent run and grounding planning in generic priors rather than the user's actual research history. \textit{(iii) current systems cannot internalize implicit preferences}. Feedback such as preferring simpler methods or wanting more efficiency analysis is too diffuse to encode as rules and too nuanced to survive compression into memory entries. Without a mechanism that converts such signals into persistent parameter-level changes, preferences fade as soon as the context window shifts.

We propose NanoResearch, a multi-agent framework that addresses these gaps through tri-level co-evolution (Figure~\ref{fig:teaser}(b)). A skill bank distills recurring operations into compact procedural rules reusable across projects, so that hard-won execution knowledge survives between runs. A memory module maintains user-bound and project-bound records that ground every planning decision in the user's actual research history rather than generic priors. A label-free policy learning mechanism converts free-form feedback into persistent parameter updates of the planner, allowing implicit preferences to reshape coordination behavior across subsequent decisions. These components are individually necessary but insufficient in isolation: procedural knowledge without user context cannot differentiate between users, contextual memory without procedural knowledge can diagnose but not prevent recurring failures, and both without preference alignment remain unable to track evolving user intent. It forms a co-evolutionary loop whereby skill execution populates memory, accumulated memory strengthens planning, and preference learning realigns the system toward each user.

Extensive experiments across 20 research topics spanning seven domains demonstrate that NanoResearch consistently outperforms existing systems under both simulated and human researcher evaluations. NanoResearch produces higher-quality research outputs while achieving stronger preference alignment, and its performance improves progressively over successive research cycles. These results suggest that personalization is not merely a desirable add-on but a fundamental axis along which autonomous research systems must evolve, and that tri-level co-evolution offers a viable path toward systems that grow more effective the longer they collaborate with a given researcher.

\section{Related Work}

\noindent\textbf{End-to-end research automation.} An emerging line of work targets end-to-end scientific automation spanning the full research lifecycle from ideation to paper writing~\cite{lu2024ai, yamada2025ai, tang2025ai, lyu2026evoscientist, weng2025deepscientist, yang2023ai, xie2025empirical}. As a pioneering effort, The AI Scientist~\cite{lu2024ai} realizes the first such fully automated pipeline, culminating in an LLM-based reviewing process, and its successor AI Scientist-v2~\cite{yamada2025ai} further incorporates agentic tree search to better explore research decisions. Other concurrent efforts~\cite{tang2025ai, lyu2026evoscientist, weng2025deepscientist, yang2023ai, xie2025empirical} instead adopt multi-agent architectures that orchestrate specialized agents to collaboratively drive the research process: EvoScientist~\cite{lyu2026evoscientist} equips agents with persistent memory and self-evolution to distill and reuse strategies from past trajectories; DeepScientist~\cite{weng2025deepscientist} formulates discovery as goal-driven Bayesian Optimization for long-horizon exploration; and AI-Researcher~\cite{tang2025ai} decomposes concepts into atomic units linking formulations to code, refined via mentor-guided agent loops. However, most existing systems still operate as static pipelines~\cite{lu2024ai, yamada2025ai, tang2025ai}, and the few attempts at dynamic adaptation~\cite{lyu2026evoscientist} remain limited to passive memory logging, failing to efficiently accumulate experience or accommodate individual user needs. In contrast, our work achieves multi-level self-evolution across skills, memory, and planner policy, and leverages user profiles together with feedback to deliver personalized outputs.

\noindent\textbf{Task-specific research automation.} Early efforts on AI scientists primarily aimed to assist human researchers in specific subtasks rather than replacing them. Even before the LLM era~\cite{touvron2023llama, bai2023qwen, guo2025deepseek, jiang2024mixtral}, prior work had explored using AI to support scientific research~\cite{lee2020biobert, cachola2020tldr, huang2019clinicalbert, beltagy2019scibert, clune2019ai}, and recent studies further leverage foundation models~\cite{team2025kimi, bai2025qwen3} to enhance assistance at individual research stages~\cite{shao2025omniscientist, team2025internagent}. Some efforts focus on literature understanding, like PaperQA~\cite{lála2023paperqaretrievalaugmentedgenerativeagent}, which answers scientific questions by retrieving and reasoning over relevant papers. Another line targets novel idea generation, with Nova~\cite{hu2024nova} retrieving external knowledge to enhance novelty and ResearchAgent~\cite{baek2025researchagent} augmenting LLMs with an entity-centric knowledge store and iterative reviewing agents. Moving from ideation to reproduction, AutoP2C~\cite{lin2025autop2c} converts papers into code via a multi-agent pipeline, while ResearchCodeAgent~\cite{gandhi2025researchcodeagent} iteratively refines an initial codebase with dynamic planning.

\section{Method}

\subsection{Overview}

\begin{figure}[t]
  \centering
  \includegraphics[width=\textwidth]{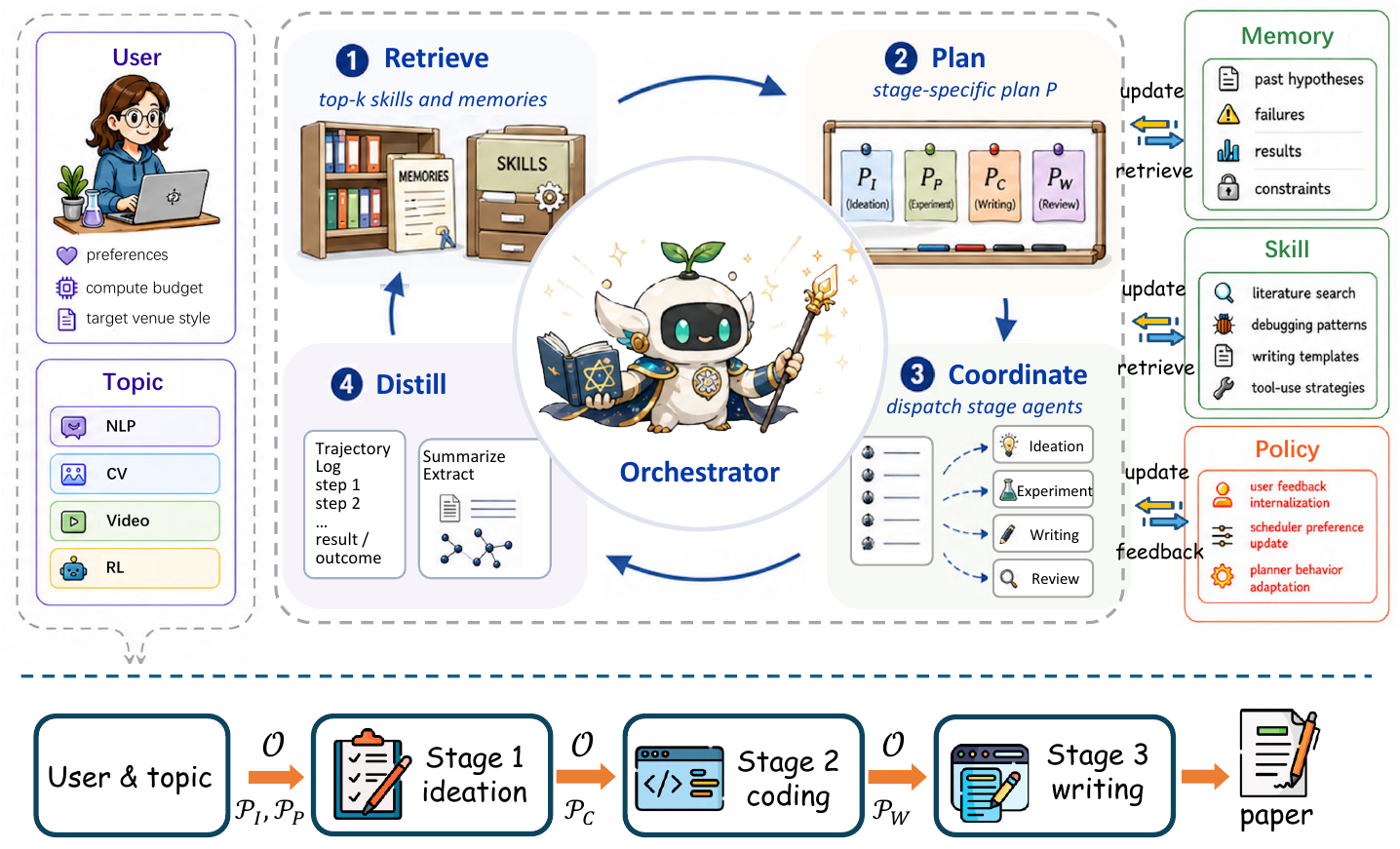}
  \caption{\textbf{The NanoResearch framework.} An Orchestrator $\mathcal{O}$ processes a personalized research request and coordinates a three-stage pipeline (ideation, experimentation, writing) to produce a publication-ready paper. A Skill Bank $\mathcal{S}$, a Memory Module $\mathcal{M}$, and policy learning jointly accumulate experience and drive self-evolution across cycles.}
  \label{main_fig}
\end{figure}

Unlike existing automated research systems~\cite{lyu2026evoscientist, tang2025ai} that follow rigid workflows, we propose \textbf{NanoResearch}, a self-evolving framework that turns a user-specified topic $\mathcal{T}$ into a complete academic paper $\mathcal{P}$. To tailor the pipeline to each researcher, the system first constructs a user profile $\mathcal{U}$ via interactive queries, serving as persistent context for all subsequent decisions. As illustrated in Figure~\ref{main_fig}, the workflow comprises three stages: (1) \textbf{Idea Generation and Planning}, (2) \textbf{Experimental Validation and Optimization}, and (3) \textbf{Paper Writing and Review}, supported by a \textbf{Skill Bank} $\mathcal{S}$ and a \textbf{Memory Module} $\mathcal{M}$, coordinated by an \textbf{Orchestrator} $\mathcal{O}$ that retrieves relevant entries before each task and updates both stores afterward. Users provide natural-language feedback $\mathcal{F}$ at the end of each stage, which $\mathcal{O}$ internalizes into its planner policy, turning explicit feedback into persistent preferences.

\subsection{NanoResearch Pipeline: A Self-Evolving Research System}

\subsubsection{Stage I: Idea Generation and Planning}

The initial stage transforms a user-specified research topic $\mathcal{T}$ into a novel, executable experiment blueprint $\mathcal{B}$, constrained by the user profile $\mathcal{U}$, through two sequential phases: \textit{Ideation} and \textit{Planning}.

\noindent\textbf{Ideation phase} begins by systematically surveying the existing literature. The Orchestrator $\mathcal{O}$ first retrieves topic- and user-aligned skills $\mathcal{S}_I \subseteq \mathcal{S}$ and memories $\mathcal{M}_I \subseteq \mathcal{M}$, and produces a high-level plan $P_I$ outlining the survey scope and hypothesis generation strategy:
\begin{equation}
    \mathcal{S}_I, \mathcal{M}_I = \text{Retrieve}(\mathcal{S}, \mathcal{M} \mid \mathcal{T}, \mathcal{U}), \quad P_{I} = \text{Plan}(\mathcal{T}, \mathcal{U} \mid \mathcal{S}_I, \mathcal{M}_I). \quad \triangleright\;\mathcal{O}
\end{equation}
Guided by $P_{I}$, the system queries academic databases (e.g., arXiv, Semantic Scholar) to retrieve relevant papers $L$, and applies a \textit{quantitative evidence extraction} mechanism that parses performance scores directly from the papers to yield grounded evidence $E$ and mitigate hallucination. A ReAct-based reasoning loop then identifies research gaps and proposes candidate hypotheses $H = \{h_1, \dots, h_K\}$, after which an automated \textit{novelty verification} step queries the databases with each $h_k$ to filter out prior-work overlaps, yielding the most promising hypothesis $h^* \in H$.

\noindent\textbf{Planning phase} translates $h^*$ into a rigorous, JSON-formatted experiment blueprint $\mathcal{B}$. The Orchestrator is invoked again to retrieve execution-level context and produce a high-level plan $P_P$:
\begin{equation}
    \mathcal{S}_P, \mathcal{M}_P = \text{Retrieve}(\mathcal{S}, \mathcal{M} \mid h^*, \mathcal{U}), \quad P_{P} = \text{Plan}(h^*, \mathcal{U} \mid \mathcal{S}_P, \mathcal{M}_P). \quad \triangleright\;\mathcal{O}
\end{equation}
Guided by $P_{P}$, $\mathcal{B}$ is instantiated with concrete specifications including datasets, baselines, proposed architecture, evaluation metrics, and ablation studies, and then undergoes an automated \textit{peer-review-like correction loop}: an internal LLM reviewer critiques $\mathcal{B}$ for infeasible designs or unfair comparisons, producing a critique $c_{\mathcal{B}}$ that drives iterative refinement:
\begin{equation}
    \mathcal{B}^{(t+1)} = \text{Refine}(\mathcal{B}^{(t)}, c_{\mathcal{B}}^{(t)}, P_{P}, E),
\end{equation}
until $\mathcal{B}$ passes review or reaches the retry limit. Finally, the Orchestrator distills new reusable skills and memories from the trajectory:
\begin{equation}
    \mathcal{S}, \mathcal{M} \leftarrow \text{Update}(\mathcal{S}, \mathcal{M} \mid h^*, \mathcal{B}, c_{\mathcal{B}}). \quad \triangleright\;\mathcal{O}
\end{equation}

\subsubsection{Stage II: Experimental Validation and Optimization}

Following the formulation of $\mathcal{B}$, this stage transitions from conceptual design to empirical validation. 

\noindent\textbf{Setup and Coding phase} first prepares the environment by cloning suitable base repositories and staging the datasets specified in $\mathcal{B}$. To align the generated code with $\mathcal{U}$, the Orchestrator $\mathcal{O}$ retrieves coding-specific skills $\mathcal{S}_C \subseteq \mathcal{S}$ and project memories $\mathcal{M}_C \subseteq \mathcal{M}$, and produces a coding plan $P_C$:
\begin{equation}
    \mathcal{S}_C, \mathcal{M}_C = \text{Retrieve}(\mathcal{S}, \mathcal{M} \mid \mathcal{B}, \mathcal{U}), \quad P_{C} = \text{Plan}(\mathcal{B}, \mathcal{U} \mid \mathcal{S}_C, \mathcal{M}_C). \quad \triangleright\;\mathcal{O}
\end{equation}
Guided by $P_C$, the Coding agent instantiates a self-contained codebase $\mathcal{W}$ comprising model definitions, training scripts, evaluation pipelines, and cluster submission scripts.

\noindent\textbf{Execution and Automated Debugging phase} deploys $\mathcal{W}$ to the target environment (e.g., a SLURM cluster). Since initial code rarely runs zero-shot, an autonomous debugging loop iteratively patches the codebase using $\mathcal{S}_C$ and $\mathcal{M}_C$ until execution succeeds or the retry budget is exhausted:
\begin{equation}
    \mathcal{W}^{(t+1)} = \text{Debug}(\mathcal{W}^{(t)} \mid \mathcal{S}_C, \mathcal{M}_C).
\end{equation}

\noindent\textbf{Analysis phase.}
Upon successful execution, raw output logs $R_{\text{raw}}$ are parsed into an analysis report $\mathcal{A}$ covering experimental results, performance comparisons, and key findings:
\begin{equation}
    \mathcal{A} = \text{Analyze}(R_{\text{raw}}, \mathcal{B}, \mathcal{T}).
\end{equation}
Finally, the Orchestrator consolidates reusable skills and memories: the experimental record, whether successful or failed, is stored in $\mathcal{M}$ with its conditions, while generalizable solutions from coding and execution are abstracted into new skills in $\mathcal{S}$:
\begin{equation}
    \mathcal{S}, \mathcal{M} \leftarrow \text{Update}(\mathcal{S}, \mathcal{M} \mid \mathcal{W}, \mathcal{A}). \quad \triangleright\;\mathcal{O}
\end{equation}

\subsubsection{Stage III: Paper Writing and Review}

The final stage integrates prior outputs into a publication-ready LaTeX manuscript.

\noindent\textbf{Writing phase.}
To maintain narrative consistency and adhere to venue-specific conventions in $\mathcal{U}$, the Orchestrator $\mathcal{O}$ retrieves writing-specific skills $\mathcal{S}_W \subseteq \mathcal{S}$ and project memories $\mathcal{M}_W \subseteq \mathcal{M}$, and formulates a structured writing plan $P_W$:
\begin{equation}
    \mathcal{S}_W, \mathcal{M}_W = \text{Retrieve}(\mathcal{S}, \mathcal{M} \mid \mathcal{B}, \mathcal{A}, \mathcal{U}), \quad P_W = \text{Plan}(\mathcal{B}, \mathcal{A}, \mathcal{U} \mid \mathcal{S}_W, \mathcal{M}_W). \quad \triangleright\;\mathcal{O}
\end{equation}
Following $P_W$, the Writing agent drafts the manuscript \textit{section-by-section} to alleviate context limitations and avoid catastrophic forgetting. 

\noindent\textbf{Review phase.}
To ensure an unbiased evaluation, the Review agent operates without the skill or memory retrieval used in earlier stages. Acting as a strict external reviewer, it critiques the draft on logical coherence, claim validity, and formatting correctness, producing targeted feedback $f_R$:
\begin{equation}
    \text{Draft}^{(t+1)} = \text{Revise}(\text{Draft}^{(t)}, f_R^{(t)}),
\end{equation}
which repeats until predefined quality thresholds are met, yielding the final paper $\mathcal{P}$. The Orchestrator then distills reusable knowledge, e.g., writing techniques and revision strategies, into $\mathcal{S}$ and $\mathcal{M}$:
\begin{equation}
    \mathcal{S}, \mathcal{M} \leftarrow \text{Update}(\mathcal{S}, \mathcal{M} \mid \mathcal{P}, f_R). \quad \triangleright\;\mathcal{O}
\end{equation}

\subsection{Foundations of Self-Evolution: Memory, Skills, and Planning}

\subsubsection{Memory and Skill Management}
The Orchestrator $\mathcal{O}$ drives the continuous evolution through the Skill Bank $\mathcal{S}$ and the Memory Module $\mathcal{M}$, relying on two core mechanisms: context-aware retrieval and trajectory-based updating.

\noindent\textbf{Retrieval Mechanism.}
Before each task, $\mathcal{O}$ retrieves the top-$k$ skills $\mathcal{S}_C$ and memories $\mathcal{M}_C$ relevant to the current context $C$ (e.g., $\mathcal{T}$, $\mathcal{U}$, $\mathcal{B}$) via a heuristic scoring function:
\begin{equation}
    \mathcal{S}_C = \underset{s \in \mathcal{S}}{\text{top-}k}\;\text{score}(C, s), \quad \mathcal{M}_C = \underset{m \in \mathcal{M}}{\text{top-}k}\;\text{score}(C, m).
\end{equation}
The score combines keyword matching, tag alignment, and recency, with weights adapted to the target: skill retrieval prioritizes usage frequency and confidence to surface robust strategies (e.g., debugging patterns), while memory retrieval enforces strict condition matching to return only project-specific experiences (e.g., prior outcomes) from comparable settings.

\noindent\textbf{Update Mechanism.}
Upon completing a stage, $\mathcal{O}$ reflects over the trajectory $\tau$ (actions, critiques, outcomes), distilling generalizable rules (e.g., debugging strategies) into the Skill Bank and project-specific experiences (e.g., failed hypotheses) into the Memory Module:
\begin{equation}
    \mathcal{S}^{(t+1)} = \mathcal{S}^{(t)} \cup \text{Distill}_{\text{skill}}(\tau), \quad \mathcal{M}^{(t+1)} = \mathcal{M}^{(t)} \cup \text{Summarize}_{\text{mem}}(\tau).
\end{equation}
To prevent unbounded growth, $\mathcal{O}$ further merges semantically overlapping entries, keeping both stores compact for future cycles.

\subsubsection{Adaptive Planning}

While $\mathcal{S}$ and $\mathcal{M}$ capture broad procedural knowledge and project facts, we further internalize fine-grained, user-specific preferences (e.g., coding style, analytical focus). At the end of each stage, the user provides immediate natural-language feedback $\mathcal{F}$, which we encode directly into the Orchestrator's \textbf{planner model} $\pi_\theta$ rather than $\mathcal{S}$ or $\mathcal{M}$, where it risks being compressed or missed at retrieval. Since $\mathcal{F}$ is free-form language rather than scalar rewards or preference pairs, we adopt Self-Distillation Policy Optimization (SDPO)~\cite{buening2026aligning}, which converts a single feedback instance into a dense, token-level learning signal without any reward model or preference annotation. Formally, given the Orchestrator's input $x$ and the planner's initial trajectory $y \sim \pi_\theta(\cdot \mid x)$, it treats the feedback-conditioned model $\pi_\theta(\cdot \mid x, \mathcal{F}, y_{<t})$ as a \textit{self-teacher} and updates the student $\pi_\theta(\cdot \mid x, y_{<t})$ to match its token distribution. Following \cite{buening2026aligning}, the SDPO gradient is a logit-level policy gradient:
\begin{equation}
\nabla_\theta \mathcal{L}_{\text{SDPO}}(\theta) = -\,\mathbb{E}_{y \sim \pi_\theta(\cdot \mid x)} \left[ \sum_{t=1}^{|y|} \mathbb{E}_{\hat{y}_t \sim \pi_\theta(\cdot \mid x, y_{<t})} \left[ A^{\text{SDPO}}_{t}(\hat{y}_t) \cdot \nabla_\theta \log \pi_\theta(\hat{y}_t \mid x, y_{<t}) \right] \right],
\end{equation}
with the dense token-level advantage estimated via the self-teacher:
\begin{equation}
    A^{\text{SDPO}}_{t}(\hat{y}_t) = \log \frac{\pi_\theta(\hat{y}_t \mid x, \mathcal{F}, y_{<t})}{\pi_\theta(\hat{y}_t \mid x, y_{<t})}.
\end{equation}
Applied after each feedback round, this update progressively internalizes user preferences into the planner's parameters, enabling NanoResearch to satisfy user preferences over successive cycles.

\section{Experiments}
\subsection{Experiment Setup}

\begin{figure}[ht]
  \centering
  \includegraphics[width=\textwidth]{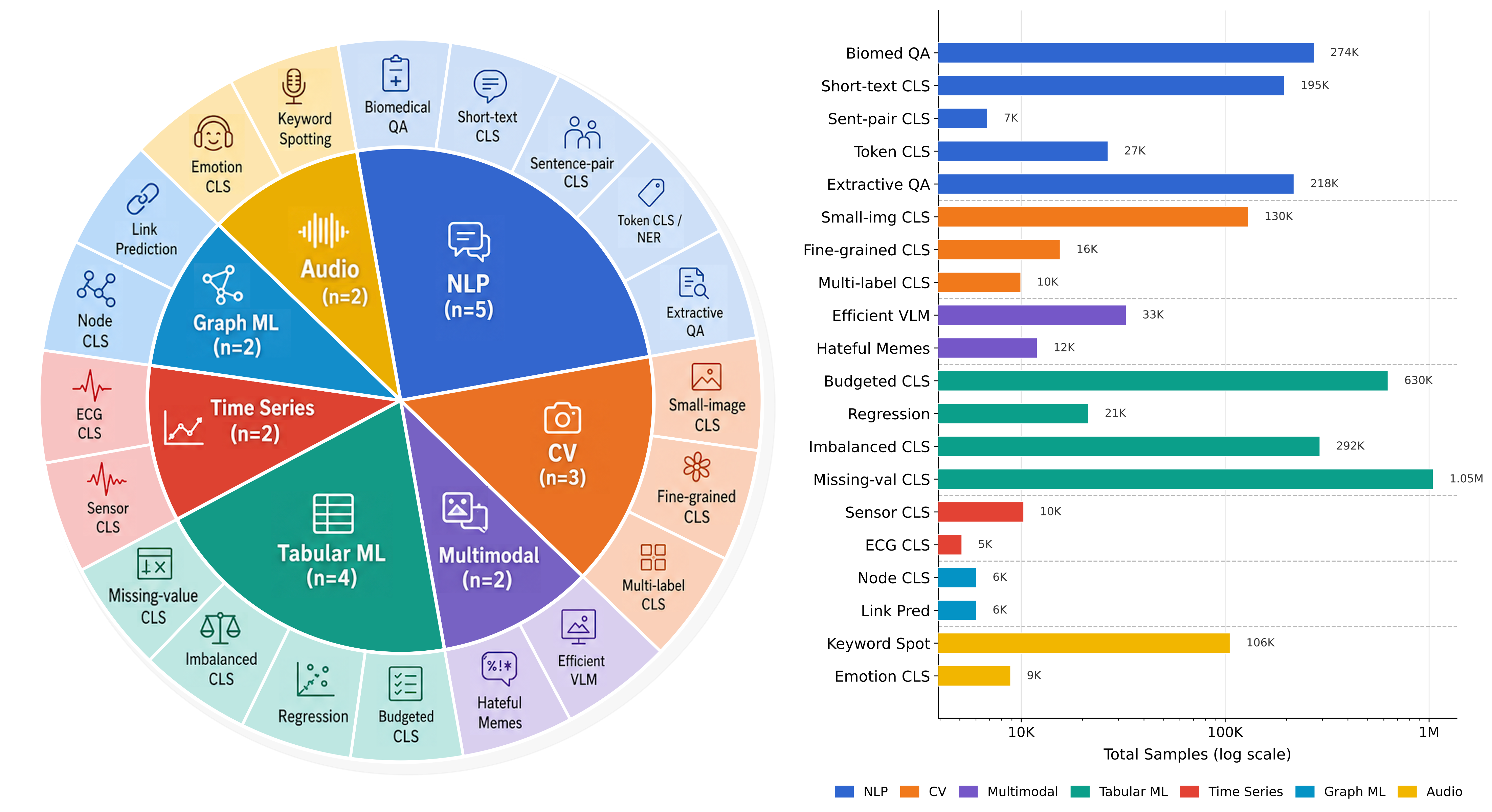}
  \caption{\textbf{Composition of our benchmark.}
  The 20 research tasks span seven domains, and cover a wide variety of subtasks (left), with dataset sizes ranging from $\sim$5K to over 1M samples (right).}
  \label{fig:dataset_composition}
\end{figure}

To comprehensively evaluate NanoResearch, we build a benchmark of 20 research tasks spanning seven domains (NLP, CV, Multimodal, Tabular ML, Time Series, Graph ML, and Audio). For each task, we construct an LLM-simulated scientist with their own preferences and constraints, who provides feedback throughout the pipeline, enabling personalized, multi-round evaluation. To assess self-evolution, we run NanoResearch for multiple rounds on each task and compare outputs across successive iterations. Details and the full task composition are provided in Section~\ref{sec:benchmark_construction} and Figure~\ref{fig:dataset_composition}.

\noindent\textbf{Baselines.} We compare NanoResearch against four representative end-to-end automated research systems: AI-Researcher~\cite{tang2025ai}, DeepScientist~\cite{weng2025deepscientist}, EvoScientist~\cite{lyu2026evoscientist}, and AI Scientist-v2~\cite{yamada2025ai}. All systems are run under the same task specifications and evaluated with identical metrics.

\noindent\textbf{Metrics.} We evaluate each system along five dimensions spanning the full research lifecycle: (1) \textit{Compliance} (Align.), how well the output matches the user's specified topic and requirements; (2) \textit{Executability} (E2E), the fraction of runs that complete the full pipeline with executable experiments and a final paper; (3) \textit{Effectiveness} (Perf.), the average task accuracy of the produced method; (4) \textit{Innovation} (Novel.), the originality of the proposed idea relative to prior work; and (5) \textit{Expression} (Writ.), the writing quality of the final paper. All subjective scores are rated by an LLM judge.

\noindent\textbf{Implementation Details.} Literature retrieval is performed via the OpenAlex API. The Planner of the Orchestrator is the only trainable component and is instantiated as Qwen3-8B. For the other agents, Ideation, Planning, and Setup/Execution use DeepSeek-V3.2; Coding/Debugging uses GPT-5.3-Codex; Writing and figure prompt/code generation use Claude Sonnet 4.6; figure image generation uses Gemini 3.1 Flash; Review uses Gemini 3.1 Flash Lite; and Revision uses Gemini 3 Pro. 

\subsection{Benchmark Construction}
\label{sec:benchmark_construction}

To support the personalized, multi-round evaluation, we construct a benchmark of 20 research tasks together with a simulated researcher for each task. The construction is fully driven by Claude, which serves both as the topic generator and as the in-the-loop user during NanoResearch runs.

\subsubsection{Construction Protocol}
\label{sec:benchmark_protocol}

We prompt Claude to role-play as 20 distinct scientists, each proposing
a concrete research topic together with the relevant contextual
information. To ensure breadth and comparability across tasks, the
generated topics provide cross-domain coverage spanning NLP, CV,
Multimodal, Tabular ML, Time Series, Graph ML, and Audio, and each
topic specifies explicit user requirements such as reproducibility
and methodological focus.

\subsubsection{Topic Schema}
\label{sec:benchmark_schema}

Each topic produced by Claude follows a fixed schema with the following fields: \texttt{question\_id}, \texttt{domain}, \texttt{difficulty}, \texttt{background}, \texttt{problem\_statement}, \texttt{baselines}, \texttt{datasets}, \texttt{user\_requirements}, and \texttt{extra\_context}. Together, these fields define a self-contained research request that captures both the scientific problem and the simulated researcher's personal preferences and constraints, providing a stable interface between the benchmark and the NanoResearch pipeline.

\subsubsection{Simulated Researcher Feedback}
\label{sec:benchmark_feedback}

Beyond topic generation, Claude continues to act as the corresponding scientist throughout each NanoResearch run. After observing the intermediate artifacts produced at each stage of the pipeline (ideation, experimentation, and writing), Claude provides feedback that is consistent with the persona's predefined preferences, constraints, and \texttt{user\_requirements}. 

\subsubsection{Role-Play Prompt}
\label{sec:benchmark_prompt}

The full role-play prompt used to instruct Claude to generate the 20
benchmark tasks is shown below. The prompt specifies the target
domains and the construction requirements.

\begin{tcolorbox}[
  breakable,
  colback=gray!5,
  colframe=black!60,
  boxrule=0.5pt,
  arc=2pt,
  left=6pt, right=6pt, top=6pt, bottom=6pt,
  title=\textbf{Role-play prompt for benchmark topic generation},
  fonttitle=\small\bfseries
]
\small
You are helping construct a benchmark suite for evaluating autonomous research agents. Generate 20 realistic research task specifications across the following seven domains: NLP, Computer Vision, Multimodal Learning, Tabular ML, Time Series, Graph ML, and Audio.

Each task should represent a concrete research topic that a real scientist or practitioner could give to a research collaborator. The goal is to test whether an autonomous research agent can propose, plan, implement, and evaluate a reproducible, benchmark-comparable research idea under explicit user constraints.

\medskip
For each task, output a JSON object with the following fields:
\begin{itemize}\setlength\itemsep{1pt}
  \item \texttt{question\_id}: a short unique identifier.
  \item \texttt{domain}: one of NLP, CV, Multimodal, Tabular ML, Time Series, Graph ML, or Audio.
  \item \texttt{difficulty}: use \texttt{"incremental\_innovation"}.
  \item \texttt{background}: a short paragraph explaining why this task is meaningful and benchmarkable.
  \item \texttt{problem\_statement}: a concrete research problem the agent should solve.
  \item \texttt{baselines}: a list of standard baselines or reference methods that should be used for comparison.
  \item \texttt{datasets}: a list of public datasets or standard benchmarks suitable for quantitative evaluation.
  \item \texttt{user\_requirements}: explicit user preferences and constraints, such as reproducibility, required ablations, metric reporting, baseline compatibility, or compute budget.
  \item \texttt{extra\_context}: additional practical constraints, such as expected resource budget, implementation feasibility, avoiding private datasets, and requiring clear ablations.
\end{itemize}

\medskip
\textbf{Construction requirements:}
\begin{enumerate}\setlength\itemsep{1pt}
  \item Cover all seven domains, with a balanced distribution across NLP, CV, Multimodal Learning, Tabular ML, Time Series, Graph ML, and Audio.
  \item Every task must be benchmarkable: it must specify public datasets, standard baselines, and evaluation-compatible outputs.
  \item Prefer tasks with clear quantitative evaluation protocols over open-ended or purely qualitative research questions.
  \item Prefer realistic incremental research problems that admit implementable methods, controlled ablations, and fair baseline comparisons.
  \item Avoid vague topics, purely theoretical topics, tasks requiring private data, and tasks whose success cannot be measured quantitatively.
  \item Ensure the topics are compatible with multiple autonomous research-agent baselines: do not rely on NanoResearch-specific mechanisms or proprietary APIs.
  \item Use a unified schema so that downstream systems can parse the topic into selected idea, proposed method, experiment plan, ablations, metrics, and benchmark targets.
\end{enumerate}

\medskip
Return exactly a JSON list of 20 task objects and no extra commentary.
\end{tcolorbox}

\subsection{Simulated Researcher Evaluation}

\subsubsection{Main Results}

Table~\ref{tab:main_results} compares NanoResearch with four representative automated research systems across 20 LLM-simulated scientists. Even in Round 1, NanoResearch surpasses all baselines on every metric and is the only system attaining a perfect 100\% end-to-end success rate, while existing systems range from 50\% (AI-Researcher, AI Scientist-v2) to 90\% (DeepScientist). We attribute this robustness to the peer-review-like blueprint correction, the autonomous debugging loop, and the dual-store retrieval that supplies relevant skills and memories at each step, which together make the pipeline resilient to the runtime errors and design flaws that often disrupt such systems.

\begin{table*}[ht]
\centering
\small
\caption{Main results across five evaluation dimensions over 20 LLM-simulated scientists. Higher is better on all metrics; \textbf{bold} denotes the best result in each column.}
\setlength{\tabcolsep}{5.2mm}
\begin{tabular}{lccccc}
    \toprule
    Method 
      & Align.\,$\uparrow$ 
      & Novel.\,$\uparrow$ 
      & E2E\,$\uparrow$ 
      & Perf.\,$\uparrow$ 
      & Writ.\,$\uparrow$ \\
    \midrule
    AI-Researcher            & 4.206 & 2.953          & 0.500          & 0.2849          & 5.402 \\
    DeepScientist            & 4.504 & 3.934          & 0.900          & 0.5634          & 4.806 \\
    EvoScientist             & 4.823 & 4.555          & 0.750          & 0.5779          & 4.953 \\
    AI Scientist-v2          & 6.656 & 3.958          & 0.500          & 0.6238          & 4.125 \\
    \midrule
    NanoResearch (Round 1)   & 8.163          & 4.960          & \textbf{1.000} & 0.6844          & 5.428 \\
    NanoResearch (Round 2)   & 8.397          & 5.164          & \textbf{1.000} & 0.7320          & 5.859 \\
    NanoResearch (Round 3)   & \textbf{8.963} & \textbf{5.645} & \textbf{1.000} & \textbf{0.7548} & \textbf{6.172} \\
    \bottomrule
  \end{tabular}
  \label{tab:main_results}
\end{table*}

\begin{wrapfigure}{r}{0.51\textwidth}
    \centering
    \includegraphics[width=\linewidth]{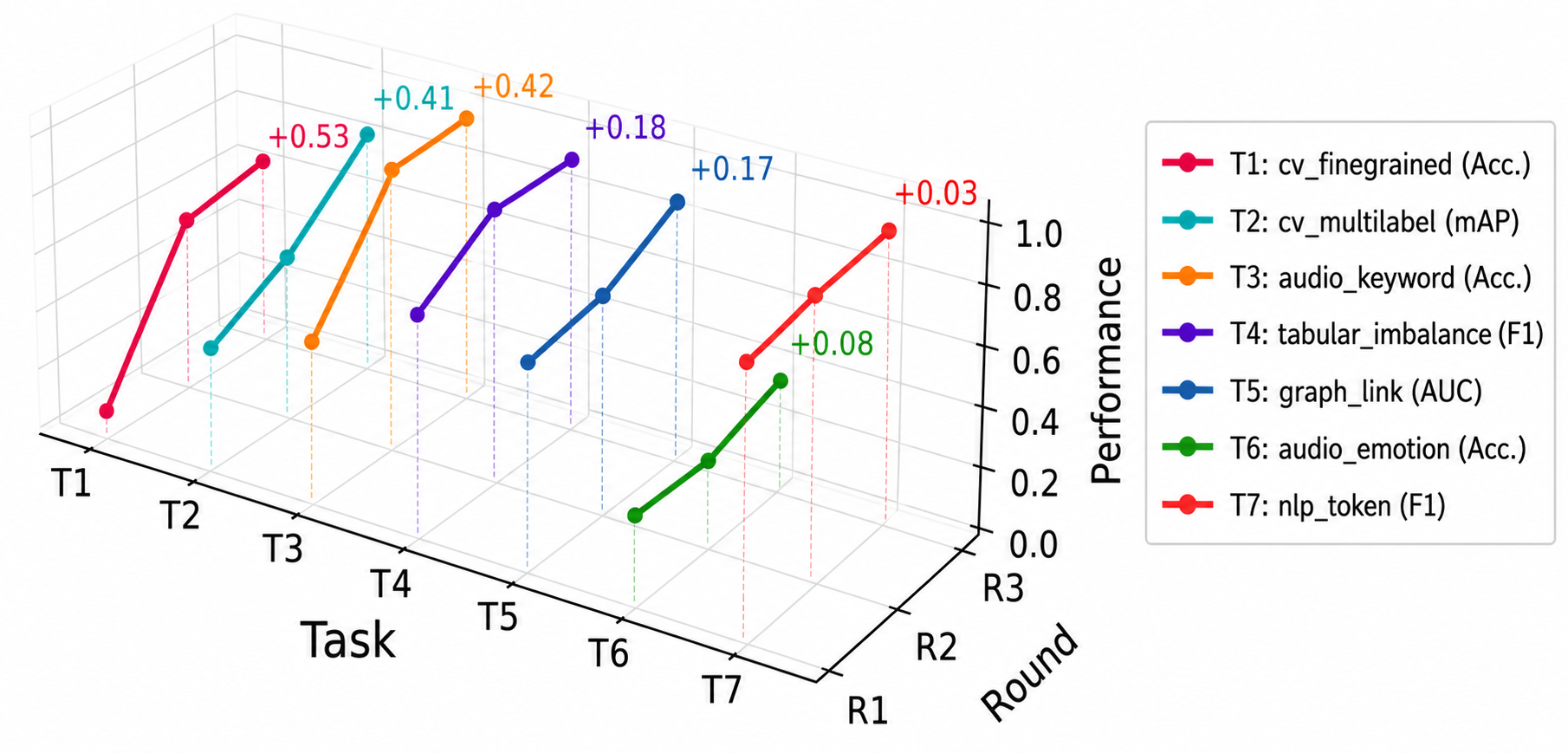}
    \caption{Per-task performance of NanoResearch.}
    \label{fig:topic_trajectories}
\end{wrapfigure}

The most pronounced advantage emerges on \textit{Compliance} (8.963 vs.\ 6.656), confirming that the user profile $\mathcal{U}$ and SDPO-based feedback internalization let NanoResearch faithfully respect heterogeneous user preferences. Performance further improves monotonically from Round 1 to Round 3 on all dimensions, with notable gains on \textit{Innovation} (4.960\,$\rightarrow$\,5.645) and \textit{Expression} (5.428\,$\rightarrow$\,6.172), showing that the Skill Bank and Memory Module help NanoResearch genuinely accumulate procedural and contextual knowledge across cycles.

\subsubsection{Ablation Studies}

We ablate each core component of NanoResearch to assess its individual
contribution. Results are summarized in Table~\ref{tab:ablation}.

\begin{table*}[ht]
  \centering
  \footnotesize
  \caption{Ablation results. \textbf{Bold} denotes the best result in
  each column.}
  \begin{tabular*}{\textwidth}{@{\extracolsep{\fill}}l|ccccc@{}}
    \toprule
    Variant
      & Align.\,$\uparrow$
      & Novel.\,$\uparrow$
      & E2E\,$\uparrow$
      & Perf.\,$\uparrow$
      & Writ.\,$\uparrow$ \\
    \midrule
    w/o Skill Bank            & 7.940 & 3.773 & 0.849 & 0.6480 & 4.75 \\
    w/o Memory                & 8.070 & 4.400 & 0.935 & 0.6590 & 5.10 \\
    w/o Planner Model         & 7.820 & 3.532 & 0.835 & 0.6420 & 4.70 \\
    w/o Preference Alignment  & 8.030 & 4.275 & 1.000 & 0.6660 & 5.05 \\
    \midrule
    Memory Only               & 7.960 & 3.899 & 0.968 & 0.6310 & 4.85 \\
    Skill Bank Only           & 7.880 & 3.715 & 0.943 & 0.6040 & 4.90 \\
    Planner + SDPO            & 7.900 & 3.860 & 0.979 & 0.6530 & 4.80 \\
    \midrule
    \textbf{NanoResearch (full)}
      & \textbf{8.163} & \textbf{4.960} & \textbf{1.000}
      & \textbf{0.6844} & \textbf{5.42} \\
    \bottomrule
  \end{tabular*}
  \label{tab:ablation}
\end{table*}

Removing the Planner Model causes the largest drop across all metrics,
confirming its central role in plan integration, while disabling the
Skill Bank lowers E2E from 1.000 to 0.849, showing that procedural
knowledge is critical for reliable execution. Removing Memory mainly
hurts novelty (4.960\,$\rightarrow$\,4.400), and removing Preference
Alignment keeps E2E at 1.000 but degrades all other dimensions,
indicating that it sharpens planning quality rather than execution
reliability. The partial configurations further reveal clear synergy:
Planner+SDPO is the strongest partial variant yet still falls short of
the full system, showing Memory, Skill Bank, and SDPO are
complementary.

\subsubsection{Efficiency Analysis}
\begin{table*}[ht]
\centering
\small
\caption{Efficiency and cost comparison across automated research systems. All values are averaged per topic. Token counts are in millions (M). GPU cost is estimated at \$2.00/hr.}
\setlength{\tabcolsep}{1.1mm}
\begin{tabular}{l|ccccc|cc}
\toprule
Method 
 & \makecell{Avg. API\\Calls $\downarrow$} 
 & \makecell{Avg. Tokens\\(M) $\downarrow$} 
 & \makecell{Avg. Runtime\\(hrs) $\downarrow$} 
 & \makecell{Avg. GPU\\Hours $\downarrow$} 
 & \makecell{API\\Cost (\$) $\downarrow$} 
 & \makecell{GPU\\Cost (\$) $\downarrow$} 
 & \makecell{Total\\Cost (\$) $\downarrow$} \\
\midrule
AI Scientist-v2            & 68.30 & 0.750 & 1.93 & 1.15 & 3.750 & 2.289 & 6.039 \\
EvoScientist               & 24.67 & 0.428 & 1.35 & 1.02 & 0.914 & 2.030 & 2.944 \\
\midrule
NanoResearch (R1)     & 23.65 & 0.117 & 2.24 & 1.75 & 0.648 & 3.509 & 4.157 \\
NanoResearch (R2)     & 18.00 & 0.092 & 1.51 & 1.13 & 0.284 & 2.258 & 2.542 \\
NanoResearch (R3)     & \textbf{15.80} & \textbf{0.073} & \textbf{1.05} & \textbf{0.60} & \textbf{0.236} & \textbf{1.194} & \textbf{1.430} \\
\bottomrule
\end{tabular}
\label{tab:cost}
\end{table*}

As shown in Table~\ref{tab:cost}, NanoResearch consistently uses far fewer tokens than the baselines, and although R1 incurs higher runtime and GPU cost due to the absence of prior skills and memories, both drop sharply in later rounds as accumulated experience helps the system converge faster. By R3, the total cost reaches only \$1.430 per topic, about 76\% lower than AI Scientist-v2 (\$6.039) and 51\% lower than EvoScientist (\$2.944), demonstrating superior research quality at lower cost with compounding efficiency gains across rounds.

\subsubsection{Skill Bank and Memory Module: Growth Across Rounds}

\begin{wraptable}{r}{0.42\textwidth}
\centering
\caption{Growth of the Skill Bank and Memory Module across rounds.}
\label{tab:skill_memory_growth}
\footnotesize
\setlength{\tabcolsep}{9pt}
\renewcommand{\arraystretch}{0.95}
\begin{tabular}{l|cc|c}
\toprule
 & \multicolumn{2}{c|}{Bank Size} & Growth \\
\cmidrule{2-4}
Round 
 & \makecell{Skill\\/Topic} 
 & \makecell{Memory\\/Topic} 
 & \makecell{New\\/Topic} \\
\midrule
R1 & 0.80          & 6.40           & 0.80 \\
R2 & 1.00          & 8.15           & 0.20 \\
R3 & \textbf{2.30} & \textbf{12.00} & \textbf{1.30} \\
\bottomrule
\end{tabular}
\end{wraptable}

To examine how self-evolution drives progressive improvement, we analyze the growth of the Skill Bank $\mathcal{S}$ and Memory Module $\mathcal{M}$ across successive rounds, as summarized in Table~\ref{tab:skill_memory_growth}. Both stores expand consistently, with the per-topic Skill Bank size growing from 0.80 to 2.30 and the Memory Module from 6.40 to 12.00 between R1 and R3. This steady accumulation indicates that the Orchestrator effectively distills reusable procedural rules and project-specific experiences from each trajectory, enabling subsequent cycles to draw on richer context and more diverse strategies, which aligns with the performance gains observed in Table~\ref{tab:main_results}.

\subsubsection{Case Study by Different Users}
\begin{figure}[ht]
  \centering
  \includegraphics[width=\textwidth]{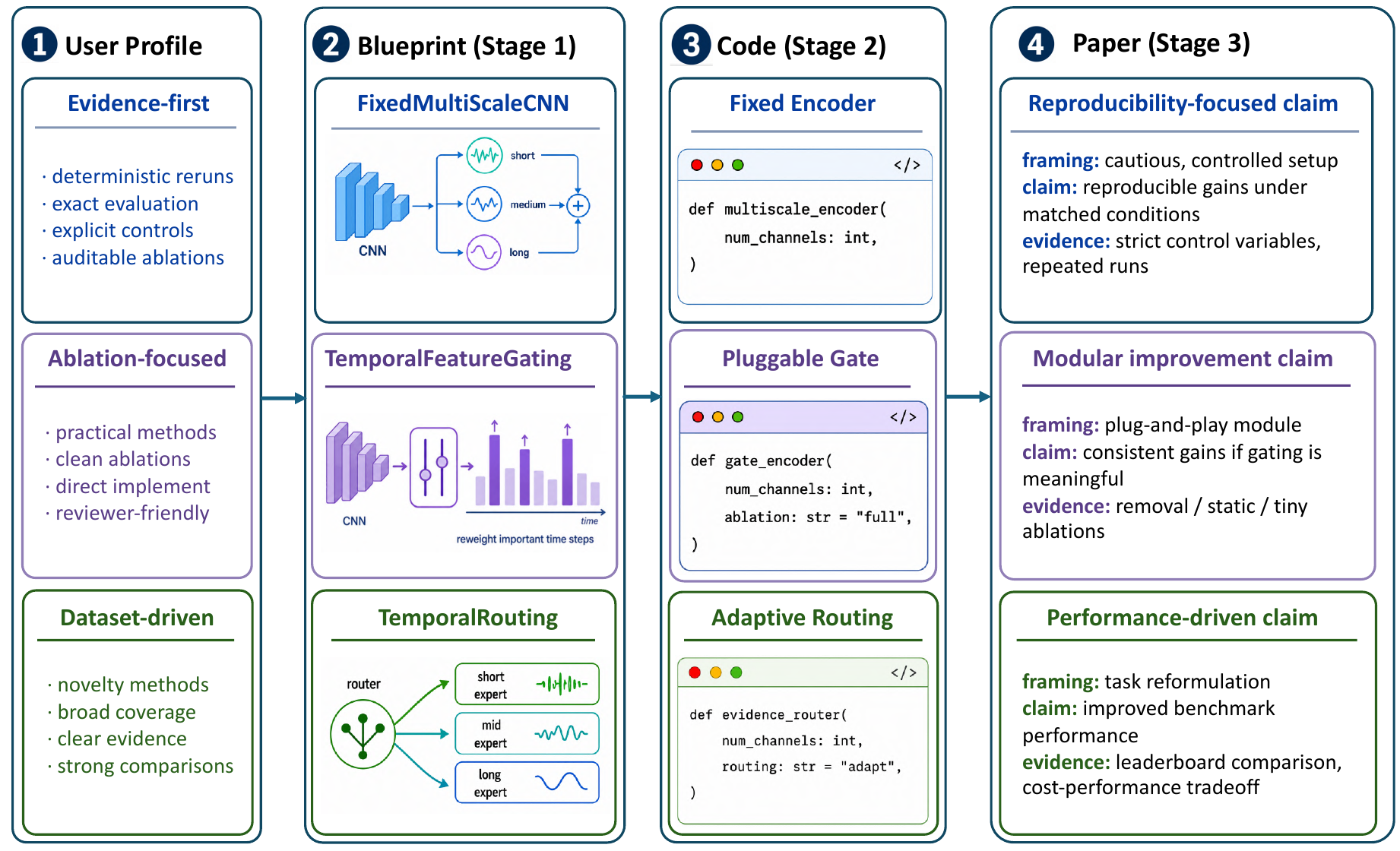}
  \caption{Case study on UCI HAR: three simulated users with
\emph{Conservative}, \emph{Practical}, and \emph{Exploratory} tastes
lead visibly different blueprints, code, and paper claims at every stage.}
  \label{fig:case_study_profiles}
\end{figure}

To probe how user profiles reshape the research trajectory beyond metrics,
we instantiate three simulated researchers with \emph{Evidence-first},
\emph{Ablation-focused}, and \emph{Dataset-driven} tastes on the same UCI HAR topic.
As shown in Figure~\ref{fig:case_study_profiles} and detailed in
Appendix~\ref{app:case-study}, the blueprint, code, and paper claim
diverge sharply across profiles. The \emph{Evidence-first} user adopts a
FixedMultiScaleCNN blueprint realized as a Fixed Encoder, framing the
paper around reproducibility-focused claims. The \emph{Ablation-focused}
user adopts TemporalFeatureGating as a Pluggable Gate, framing the paper
around modular improvement claims. The \emph{Dataset-driven} user adopts
TemporalRouting as Adaptive Routing, framing the paper around
performance-driven claims. Across all three runs, the user profile $\mathcal{U}$ shapes every stage
consistently, producing outputs that differ in research \emph{taste}
rather than by chance.
\begin{table*}[!t]
  \centering
  \small
  \setlength{\tabcolsep}{5.1mm}
  \caption{Human researcher evaluation results, averaged over three
  real research tasks rated by three PhD researchers. Higher is better;
  \textbf{bold} denotes the best in each column.}
  \begin{tabular}{l|ccccc}
    \toprule
    Method
      & Align.\,$\uparrow$
      & Novel.\,$\uparrow$
      & E2E\,$\uparrow$
      & Perf.\,$\uparrow$
      & Writ.\,$\uparrow$ \\
    \midrule
    AI-Researcher           & 4.333 & 3.333 & 1.000 & 0.5495 & 4.667 \\
    AI Scientist-v2         & 5.333 & 4.000 & 1.000 & 0.4965 & 4.333 \\
    EvoScientist            & 6.000 & 4.667 & 1.000 & 0.6537 & 4.000 \\
    DeepScientist           & 6.333 & 5.000 & 1.000 & 0.6094 & 5.333 \\
    \midrule
    NanoResearch (Round 1)  & \textbf{9.333} & 6.000 & \textbf{1.000} & 0.6466 & 7.000 \\
    NanoResearch (Round 2)  & \textbf{9.333} & \textbf{7.000} & \textbf{1.000} & 0.8502 & \textbf{8.000} \\
    NanoResearch (Round 3)  & \textbf{9.333} & 6.667 & \textbf{1.000} & \textbf{0.8603} & 7.667 \\
    \bottomrule
  \end{tabular}
  \label{tab:user_study}
\end{table*}

\begin{table*}[!t]
  \centering
  \small
  \setlength{\tabcolsep}{5.2mm}
  \caption{Per-expert human evaluation scores of \textbf{NanoResearch}
  across three self-evolution rounds. Higher is better on all metrics.}
  \label{tab:human_expert_full}
  \begin{tabular}{l|c|ccccc}
    \toprule
    Expert & Round
      & Align.\,$\uparrow$
      & Novel.\,$\uparrow$
      & E2E\,$\uparrow$
      & Perf.\,$\uparrow$
      & Writ.\,$\uparrow$ \\
    \midrule
    \multirow{3}{*}{Expert~1}
      & R1 & 10 & 6 & 1 & 0.4061 & 7 \\
      & R2 & 9  & 7 & 1 & 0.7656 & 8 \\
      & R3 & 10 & 7 & 1 & 0.7871 & 8 \\
    \midrule
    \multirow{3}{*}{Expert~2}
      & R1 & 9 & 6 & 1 & 0.6760 & 7 \\
      & R2 & 9 & 7 & 1 & 0.9087 & 8 \\
      & R3 & 9 & 7 & 1 & 0.9248 & 8 \\
    \midrule
    \multirow{3}{*}{Expert~3}
      & R1 & 9  & 6 & 1 & 0.8167 & 7 \\
      & R2 & 10 & 7 & 1 & 0.8205 & 8 \\
      & R3 & 9  & 6 & 1 & 0.8325 & 7 \\
    \bottomrule
  \end{tabular}
\end{table*}

\subsection{Human Researcher Evaluation}

To validate our findings beyond LLM-simulated evaluation, we invite
three PhD researchers to run NanoResearch and the four baselines on
their own research tasks, and rate the outputs under the same
five-dimensional rubric. As shown in Table~\ref{tab:user_study},
NanoResearch dominates all baselines on every dimension already in
Round 1, and project performance improves monotonically from 0.6466
(R1) to 0.8603 (R3), confirming that the gains observed under
simulated scientists transfer to real domain experts and that
self-evolution yields tangible quality improvements. Novelty and
writing quality rise sharply from R1 to R2 but dip slightly in R3,
likely because these dimensions are most sensitive to individual taste
and stylistic preference. Nevertheless, both remain well above the R1
level and far exceed the best baseline, leaving the overall trend
clearly positive. The full per-expert breakdown of NanoResearch across
the three self-evolution rounds is provided in
Table~\ref{tab:human_expert_full}.
\begin{figure}[htbp]
    \centering
    \includegraphics[width=\linewidth]{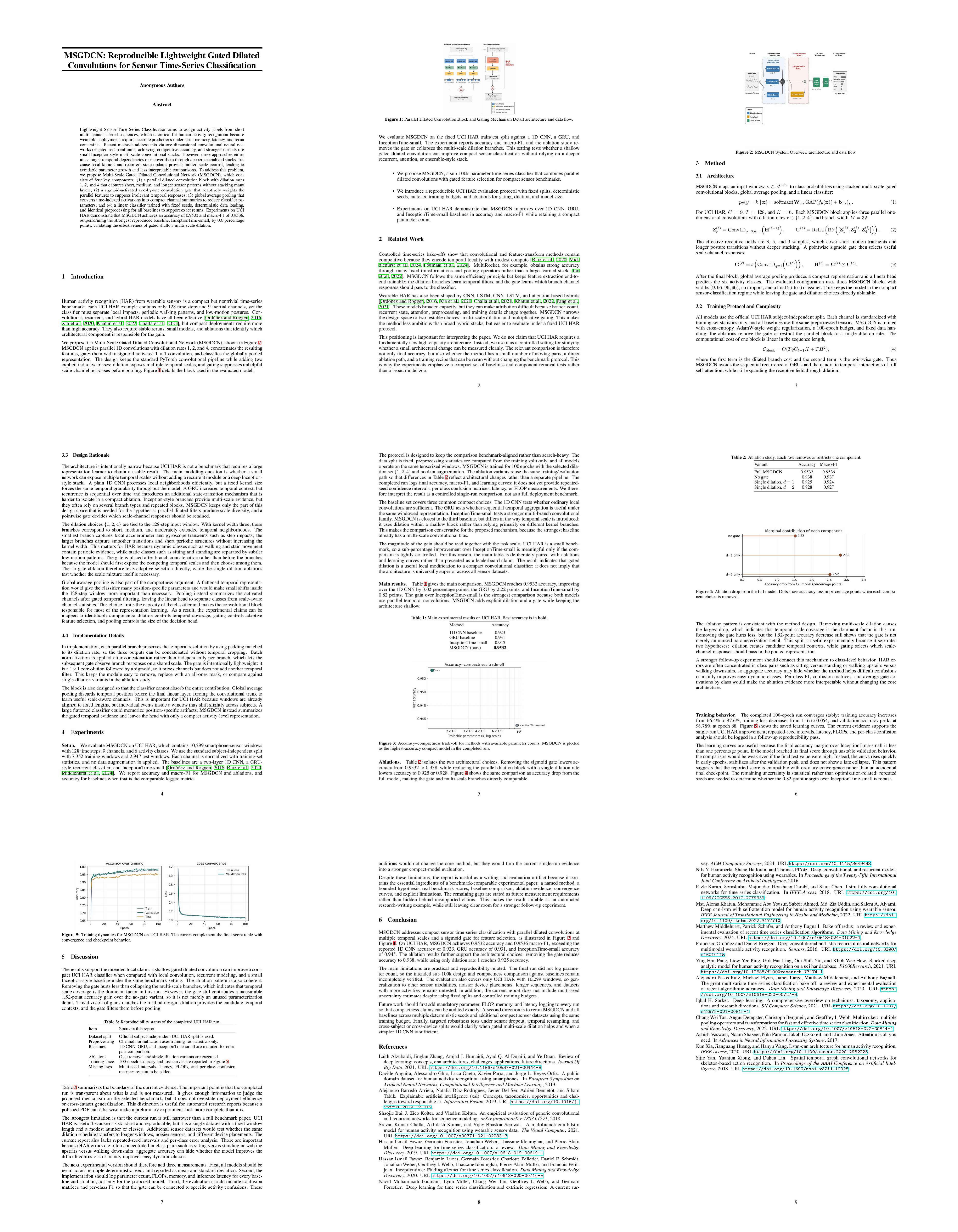}
    \caption{All pages of the system-generated sensor time-series paper
    \emph{MSGDCN: Reproducible Lightweight Gated Dilated Convolutions
    for Sensor Time-Series Classification}.}
    \label{fig:real_paper_timeseries}
\end{figure}

\begin{figure}[htbp]
    \centering
    \includegraphics[width=\linewidth]{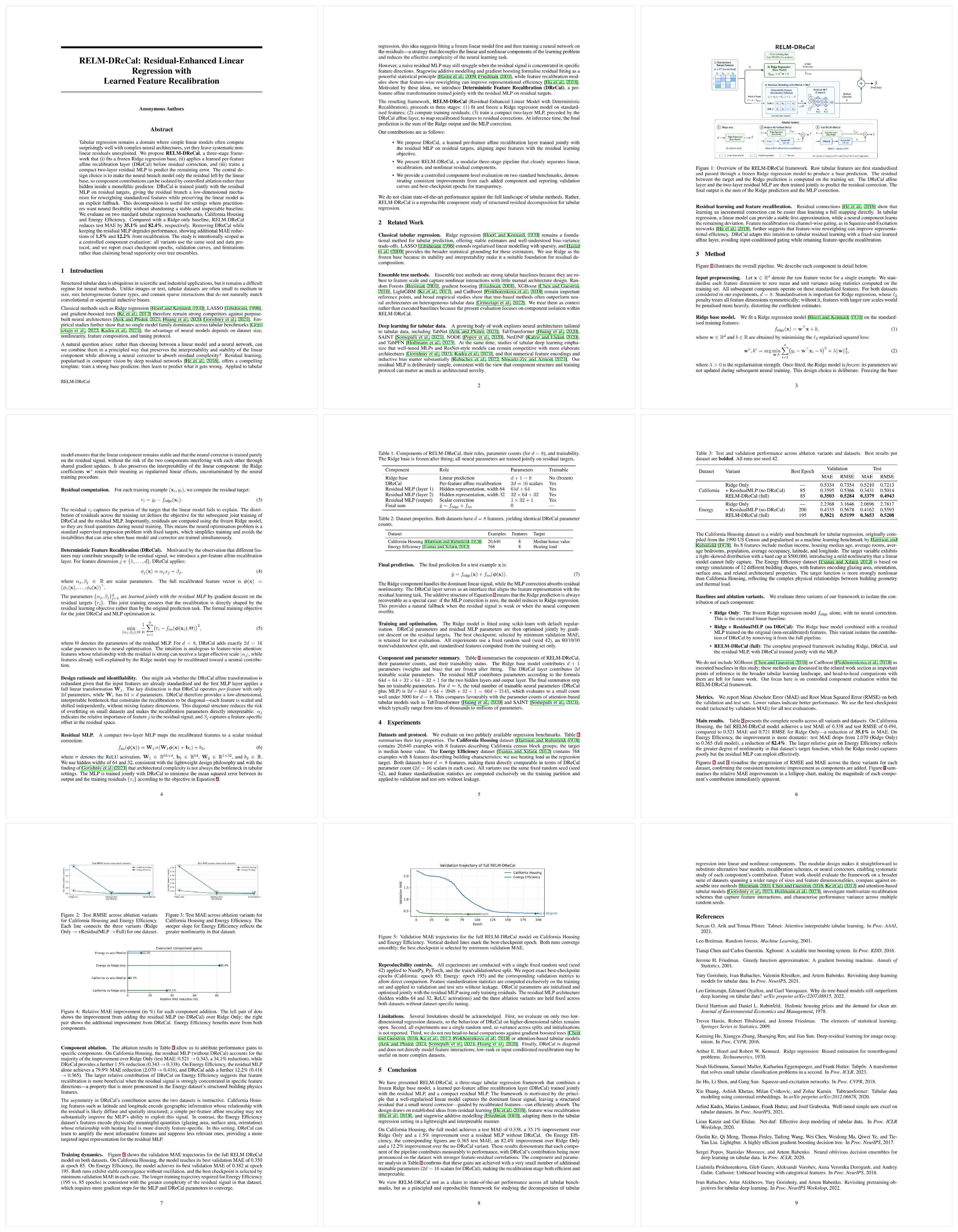}
    \caption{All pages of the system-generated tabular regression paper
    \emph{RELM-DReCal: Residual-Enhanced Linear Regression with Learned
    Feature Recalibration}.}
    \label{fig:real_paper_tabular}
\end{figure}

\begin{figure}[htbp]
    \centering
    \includegraphics[width=\linewidth]{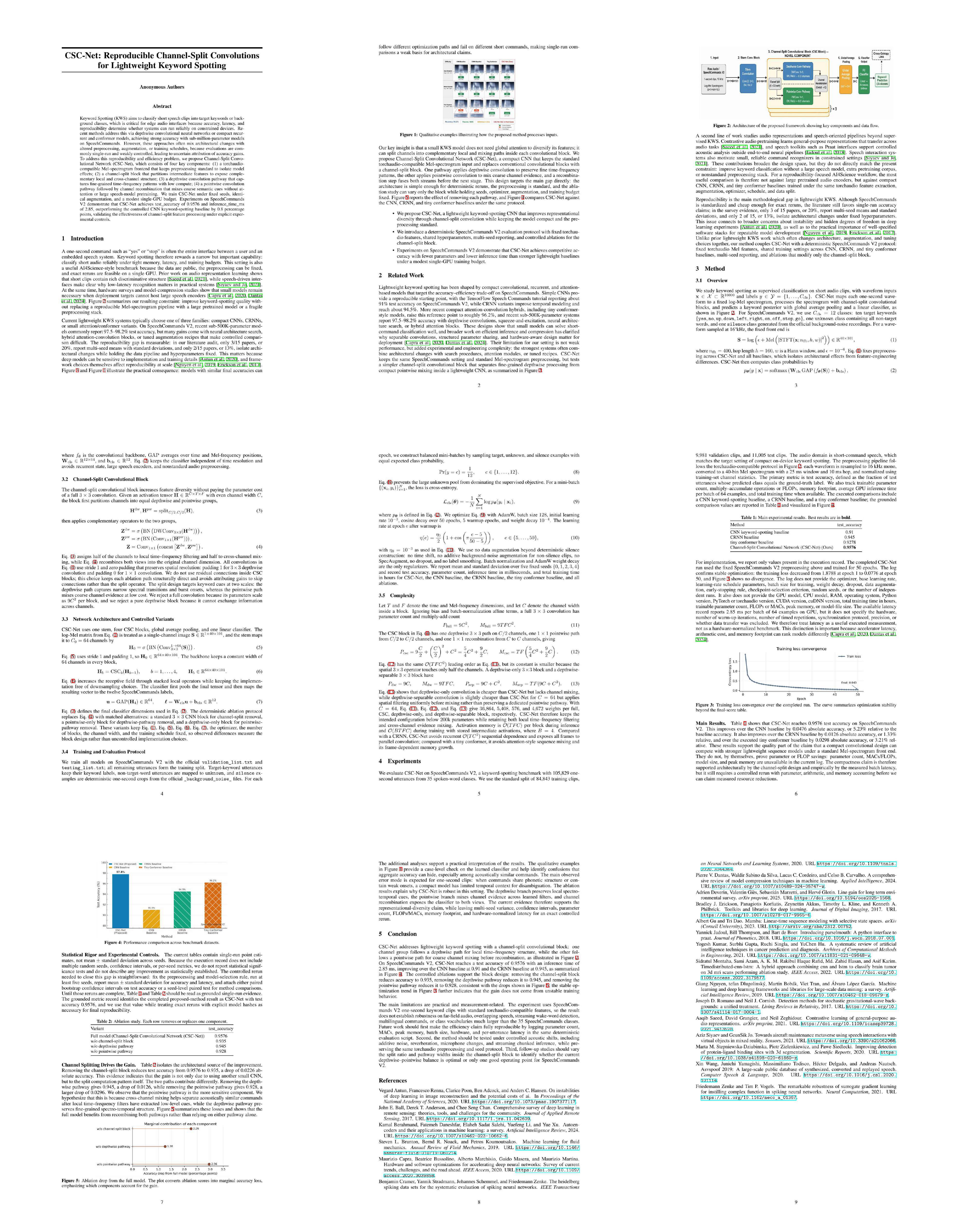}
    \caption{All pages of the system-generated keyword spotting paper
    \emph{CSC-Net: Reproducible Channel-Split Convolutions for
    Lightweight Keyword Spotting}.}
    \label{fig:real_paper_audio}
\end{figure}

\section{Real Papers Produced by Our System}
\label{sec:real_paper}

To complement the quantitative evaluations, we include three complete
papers that were actually produced end-to-end by our system, shown in
Figures~\ref{fig:real_paper_timeseries},
\ref{fig:real_paper_tabular}, and~\ref{fig:real_paper_audio}. They
demonstrate that the pipeline can deliver coherent manuscripts rather
than only isolated artefacts such as ideas, code snippets, or tables.
The three examples span distinct research domains, covering sensor
time-series classification, tabular regression, and audio keyword
spotting, illustrating that the system generalises across problem
settings rather than being tied to a single task type. Each manuscript
contains a full set of components expected of a conference submission,
including a problem motivation, related-work discussion, formal method
description with equations, an experimental protocol with baselines
and ablations, figures and tables reporting quantitative results, and
a complete reference list.



\section{Conclusion}

This work identifies personalization as a fundamental yet overlooked axis in research automation. A single, fixed pipeline cannot accommodate the diversity of researcher preferences, domain conventions, and evolving intent that characterize real scientific practice. We further show that effective personalization is not achievable through any single mechanism: it requires simultaneous adaptation at the procedural level (\textit{how tasks are executed}), the contextual level (\textit{what prior experience is retained}), and the preference level (\textit{which implicit objectives guide planning}). Our ablation results confirm that no single level alone achieves reliable alignment; the three levels are complementary, and their co-evolution yields the strongest adaptation across successive cycles. We propose NanoResearch, a tri-level co-evolutionary framework that implements layered governance across these dimensions. Experiments across 20 tasks in seven domains, evaluated by both simulated and human researchers, confirm that it consistently outperforms existing systems on all dimensions, with compounding gains across rounds and over 65\% cost reduction by the third cycle. These results suggest that personalization-aware design is not a peripheral enhancement but a prerequisite for research automation systems that are both trustworthy and practically useful.

\section{Limitations and Broader Impact}

\noindent\textbf{Limitations.} Our evaluation centers on AI/ML tasks, where research outputs can be fully realized through code and text. Extending NanoResearch to other scientific disciplines, such as biology, chemistry, or physics, where research often involves physical experimentation and instrument control, remains an important and non-trivial direction for future work.

\noindent\textbf{Broader Impact.} NanoResearch lowers the barrier to 
automated research by adapting to individual preferences and reusing 
accumulated knowledge across cycles, enabling researchers to iterate on 
ideas more efficiently. As with other multi-agent systems, it relies on 
large foundation models as its backbone, which introduces some 
computational and API costs that may limit accessibility for 
resource-constrained researchers.
\clearpage
\newpage
\bibliographystyle{plainnat}
\setcitestyle{numbers}
\bibliography{ref}

\newpage
\beginappendix

\section{User Requirement Alignment Prompt}
\label{app:alignment_prompt}

The \textit{Compliance Score} (Align.) measures how well the generated research artifacts match the user's stated requirements, including the target task, datasets, baselines, ablations, compute budget, and methodological preferences. We prompt an LLM judge with a structured JSON object describing the research task specification, the user requirements, the generated idea or selected hypothesis, the generated experiment plan or blueprint, and benchmark or execution status when available. The judge is instructed to reward outputs that are practical, benchmarkable, feasible, and aligned with the user's stated preferences, and to penalize plans that miss important requirements, use incompatible datasets or baselines, propose infeasible methods, lack required ablations, or fail to address the stated task. Scores follow a 1--10 rubric: 1--2 indicates the output largely ignores the user requirements, uses incompatible datasets or baselines, fails to address the stated task, or has no benchmark-comparable result when benchmark comparability is required; 3--4 the output is loosely related to the task but misses major requirements such as the target dataset, required baselines, feasibility constraints, compute constraints, ablation design, or benchmark comparability; 5--6 the output addresses the main task but the idea or experiment plan is incomplete, underspecified, only partially benchmark-compatible, or misses several important user preferences; 7--8 the output satisfies most user requirements, with a relevant idea, a mostly feasible and benchmarkable plan, and the required datasets, baselines, metrics, or ablations mostly covered with only minor omissions; and 9--10 the output strongly satisfies the user requirements, with the idea and experiment plan well aligned with the requested method style, appropriate datasets and baselines, feasible compute assumptions, clear evaluation metrics and ablations, and benchmark-comparable execution when required. The judge returns a structured JSON response containing the assigned \texttt{alignment\_score} and a free-text \texttt{feedback}. The full prompt template is shown below.

\begin{tcolorbox}[
  breakable,
  colback=gray!5,
  colframe=black!60,
  boxrule=0.5pt,
  arc=2pt,
  left=6pt, right=6pt, top=6pt, bottom=6pt,
  title=\textbf{Prompt: User Requirement Alignment Judge},
  fonttitle=\small\bfseries
]
\small
You are an expert research evaluator. Determine whether the generated idea and experiment plan satisfy the stated user requirements. The input is a JSON object containing: the research task specification, including domain, background, problem statement, target datasets, and known baselines; the user requirements, including preferences about feasibility, reproducibility, benchmarkability, ablations, compute budget, or method style; the generated research idea or selected hypothesis; the generated experiment plan or blueprint, including proposed method, metrics, ablations, and compute requirements when available; and benchmark or execution status when available, including whether the result is benchmark-comparable. Return JSON only with keys \texttt{alignment\_score} (a numeric score from 1 to 10) and \texttt{feedback} (a concise explanation).

\medskip
\textbf{Scoring rubric:}
\begin{itemize}\setlength\itemsep{1pt}
  \item 1--2 = the output largely ignores the user requirements, uses incompatible datasets or baselines, fails to address the stated task, or has no benchmark-comparable result when benchmark comparability is required;
  \item 3--4 = the output is loosely related to the task, but misses major requirements such as the target dataset, required baselines, feasibility constraints, compute constraints, ablation design, or benchmark comparability;
  \item 5--6 = the output addresses the main task, but the idea or experiment plan is incomplete, underspecified, only partially benchmark-compatible, or misses several important user preferences;
  \item 7--8 = the output satisfies most user requirements, with a relevant idea, a mostly feasible and benchmarkable plan, and the required datasets, baselines, metrics, or ablations mostly covered, with only minor omissions or underspecified details;
  \item 9--10 = the output strongly satisfies the user requirements, with the idea and experiment plan well aligned with the requested method style, appropriate datasets and baselines, feasible compute assumptions, clear evaluation metrics and ablations, and benchmark-comparable when required.
\end{itemize}

\medskip
Score higher when the plan is practical, benchmarkable, feasible, and aligned with the user's stated preferences. Score lower when the plan misses important requirements, uses incompatible datasets or baselines, proposes an infeasible method, lacks required ablations, has benchmark-misaligned execution, or does not address the stated task.
\end{tcolorbox}

\section{Novelty Evaluation Prompt}
\label{app:novelty_prompt}

To assess the \textit{Novelty Score} reported in the main experiments, we prompt an LLM judge with the proposed idea. The judge is instructed to focus on the \emph{core mechanism} rather than surface-level complexity, and to penalize trivial modifications such as backbone swaps, hyper-parameter tuning, regularization tricks, or data-augmentation changes. Scores follow a 1--10 rubric: 1--2 indicates a near-duplicate of existing baselines with only superficial differences; 3--4 a weak incremental modification with high overlap in core method and contribution; 5--6 moderate incremental novelty with one clear local change such as a new module, loss function, training strategy, or recombination of known components; 7--8 clearly recognizable novelty with a substantively different mechanism, method structure, or contribution logic; and 9--10 strong novelty with a non-trivial and clearly distinct core idea beyond routine recombination. The judge returns a structured JSON response containing the assigned \texttt{novelty\_score}, the \texttt{closest\_baseline} among the provided references, and a free-text \texttt{rationale}. The full prompt template is shown below.

\begin{tcolorbox}[
  breakable,
  colback=gray!5,
  colframe=black!60,
  boxrule=0.5pt,
  arc=2pt,
  left=6pt, right=6pt, top=6pt, bottom=6pt,
  title=\textbf{Prompt: Novelty Judge},
  fonttitle=\small\bfseries
]
\small
You are an expert research evaluator. Score the novelty of the proposed idea relative to the provided baselines on a 1--10 scale.

\medskip
\textbf{Scoring rubric:}
\begin{itemize}\setlength\itemsep{1pt}
  \item 1--2 = near-duplicate of the baselines with only superficial wording, hyperparameter, or training-detail changes;
  \item 3--4 = weak incremental modification with high overlap in core method and contribution;
  \item 5--6 = moderate incremental novelty with one clear local change such as a new module, loss, training strategy, or recombination of known components;
  \item 7--8 = clearly recognizable novelty with a substantively different mechanism, method structure, or contribution logic relative to the baselines;
  \item 9--10 = strong novelty with a non-trivial and clearly distinct core idea, not just module swapping or routine recombination.
\end{itemize}

\medskip
Judge primarily against the provided baselines, focus on the core mechanism rather than surface complexity, and do not over-score backbone swaps, tuning, regularization, or data augmentation. Return JSON only with keys \texttt{novelty\_score}, \texttt{closest\_baseline}, \texttt{rationale}.
\end{tcolorbox}

\section{Writing Quality Evaluation Prompt}
\label{app:writing_prompt}

The \textit{Overall Writing Quality} score and its sub-dimensions (\textit{Fluency}, \textit{Motivation Clarity}, and \textit{Preference Alignment}) are obtained by prompting an LLM judge with the full paper draft. The judge evaluates readability, organization, motivation clarity, scientific tone, and consistency with standard academic writing conventions. Scores follow a 1--10 rubric: 1--2 denotes very poor scientific writing that is hard to follow, badly structured, and not usable as a paper draft; 3--4 weak writing quality where some content is present but clarity, organization, and academic style are substantially below standard; 5--6 acceptable draft quality that is readable and partially structured but still rough, uneven, or underdeveloped; 7--8 strong research writing that is clear, coherent, and mostly polished with only minor weaknesses; and 9--10 excellent paper-quality writing that is polished, well-structured, academically credible, and close to submission quality. The judge is instructed to be strict and to use the full scale. It returns a structured JSON response containing the assigned \texttt{writing\_quality} score and a free-text \texttt{rationale}. The full prompt template is shown below.

\begin{tcolorbox}[
  breakable,
  colback=gray!5,
  colframe=black!60,
  boxrule=0.5pt,
  arc=2pt,
  left=6pt, right=6pt, top=6pt, bottom=6pt,
  title=\textbf{Prompt: Writing Quality Judge},
  fonttitle=\small\bfseries
]
\small
You are an expert evaluator of scientific writing. Read the provided paper draft and assign a single Writing Quality score from 1 to 10.

\medskip
\textbf{Scoring rubric:}
\begin{itemize}\setlength\itemsep{1pt}
  \item 1--2 = very poor scientific writing; hard to follow, badly structured, and not usable as a paper draft;
  \item 3--4 = weak writing quality; some content is present, but clarity, organization, and academic style are substantially below standard;
  \item 5--6 = acceptable draft quality; readable and partially structured, but still rough, uneven, or underdeveloped;
  \item 7--8 = strong research writing quality; clear, coherent, and mostly polished, with only minor weaknesses;
  \item 9--10 = excellent paper-quality writing; polished, well-structured, academically credible, and close to submission quality.
\end{itemize}

\medskip
Judge the score based on the overall writing quality of the full text, considering readability, organization, motivation clarity, scientific tone, and consistency with the requested writing style. Be strict. Use the full scale. Return JSON only with keys \texttt{writing\_quality}, \texttt{rationale}.
\end{tcolorbox}

\definecolor{jsonBg}{RGB}{248,248,248}
\definecolor{jsonStr}{RGB}{163,21,21}
\definecolor{jsonPunct}{RGB}{80,80,80}
\definecolor{jsonRule}{RGB}{200,200,200}

\lstdefinelanguage{json}{
    basicstyle=\ttfamily\scriptsize,
    backgroundcolor=\color{jsonBg},
    frame=single,
    framesep=4pt,
    rulecolor=\color{jsonRule},
    breaklines=true,
    breakatwhitespace=true,
    columns=fullflexible,
    keepspaces=true,
    showstringspaces=false,
    numbers=left,
    numberstyle=\tiny\color{gray},
    numbersep=6pt,
    stepnumber=1,
    tabsize=2,
    string=[s]{"}{"},
    stringstyle=\color{jsonStr},
    literate=
        *{:}{{{\color{jsonPunct}{:}}}}{1}
        {,}{{{\color{jsonPunct}{,}}}}{1}
        {\{}{{{\color{jsonPunct}{\{}}}}{1}
        {\}}{{{\color{jsonPunct}{\}}}}}{1}
        {[}{{{\color{jsonPunct}{[}}}}{1}
        {]}{{{\color{jsonPunct}{]}}}}{1},
}


\begin{figure}[!t]
  \centering
  \includegraphics[width=\textwidth]{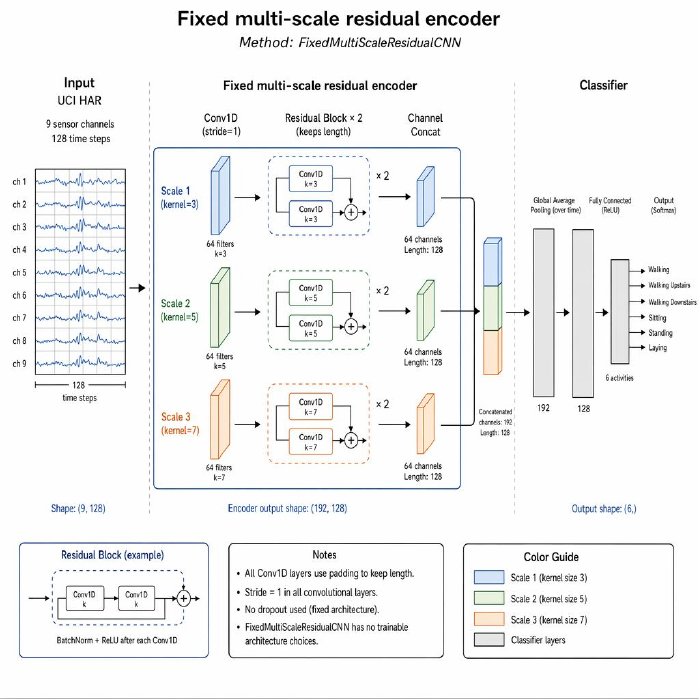}
  \caption{Architecture diagram for Profile A (Evidence-First Scientist).}
  \label{fig:case_study_profile_a}
\end{figure}



\section{Case Study: Three Profiles, One Topic}
\label{app:case-study}

This section presents an end-to-end walkthrough of how Nano Research adapts
the same research topic to three different user profiles. For each profile,
we show some intermediate pipeline outputs: the \emph{experiment blueprint}
(Stage~I), the \emph{coding output} (Stage~II), and the \emph{writing
style} (Stage~III). The shared topic and baselines are held fixed; what
varies is the inductive bias of the proposed method, the shape of the
code interface, and the tone of the writing.

\subsection{Shared Topic and Three Inductive Biases}
\label{app:case-study:shared}

\paragraph{Shared research topic.}
The shared topic is \emph{lightweight time-series sensor classification on
UCI HAR}. The goal is to design a compact model for wearable sensor
activity recognition, compare it with a 1D CNN, a GRU, and
InceptionTime-small, and keep the full experiment feasible on a single
GPU.

\paragraph{Three inductive biases.}
At a high level, the three profiles lead to three different inductive
biases over the same UCI HAR input, summarized in
Table~\ref{tab:case-study-bias}.

\begin{table}[h]
\centering
\small
\begin{tabular}{p{0.26\linewidth} p{0.28\linewidth} p{0.36\linewidth}}
\toprule
\textbf{Profile} & \textbf{Core modeling bias} & \textbf{Intuition} \\
\midrule
A: Evidence-First Scientist &
Fixed multi-scale temporal features &
Use a stable, low-variance model to test whether fixed temporal
receptive fields help. \\
B: Ablation-Focused Researcher &
Small temporal gating module &
Add one compact, inspectable contribution that can be directly removed or
simplified. \\
C: Benchmark-Driven Exploratory Researcher &
Sample-adaptive evidence routing &
Use a stronger dynamic mechanism that can support broader benchmark-facing
claims. \\
\bottomrule
\end{tabular}
\caption{Three inductive biases derived from the same shared topic. The
scientific contract is fixed; the modeling bias, code interface, and
writing style adapt to the user profile.}
\label{tab:case-study-bias}
\end{table}

\subsection{Profile A: Evidence-First Scientist}
\label{app:case-study:profile-a}

\paragraph{Stage I: Experiment blueprint.}
\begin{itemize}\setlength\itemsep{1pt}
  \item \textbf{title:} Lightweight Fixed Multi-Scale Residual CNN for
  Controlled Sensor Classification on UCI HAR.
  \item \textbf{proposed\_method.name:} FixedMultiScaleResidualCNN.
  \item \textbf{proposed\_method.description:} A standard 1D CNN is
  augmented with parallel fixed-kernel temporal branches at scales 3, 5,
  and 7, followed by feature fusion and a residual projection. The design
  keeps the model close to the baseline while testing whether broader
  fixed receptive fields improve wearable sensor classification.
  \item \textbf{proposed\_method.key\_components:} fixed multi-scale
  Conv1d branches; residual feature fusion; compact classification head.
  \item \textbf{proposed\_method.architecture:} Input (9 channels $\times$
  128 time steps) $\rightarrow$ parallel Conv1d branches with kernel sizes
  [3, 5, 7] $\rightarrow$ concatenation $\rightarrow$ 1$\times$1
  projection $\rightarrow$ residual addition $\rightarrow$ global average
  pooling $\rightarrow$ linear classifier.
  \item \textbf{ablation\_groups:} kernel-scale ablation with
  \texttt{kernel\_sizes=[5]} versus \texttt{[3,5,7]}; residual ablation
  with the skip connection removed; parameter-control ablation with
  matched channel width.
\end{itemize}

\paragraph{Stage II: Coding output.}
The code is as follows.

\begin{lstlisting}[language=Python,
                   basicstyle=\ttfamily\small,
                   columns=fullflexible,
                   keepspaces=true,
                   showspaces=false,
                   showstringspaces=false,
                   breaklines=true]
def build_fixed_multiscale_encoder(num_channels: int):
    """
    Stable encoder for a controlled UCI HAR study.
    The design uses fixed temporal scales rather than sample-adaptive routing.
    """
    return FixedScaleTemporalEncoder(
        input_channels=num_channels,
        kernel_sizes=(3, 5, 7),
        merge="concat_then_1x1",
        residual=True,
        learned_gate=False,
        learned_router=False,
    )
\end{lstlisting}

\paragraph{Stage III: Writing style.}
\begin{quote}
\itshape
Rather than claiming a new general architecture, we ask a narrower
question: do fixed multi-scale temporal features yield reproducible gains
under matched training conditions? The proposed encoder changes only the
temporal receptive-field structure of a standard 1D CNN, allowing us to
test whether short-, medium-, and longer-range convolutional features
provide consistent gains without introducing sample-adaptive routing.
\end{quote}

\subsection{Profile B: Ablation-Focused Researcher}
\label{app:case-study:profile-b}
\begin{figure}[!t]
  \centering
  \includegraphics[width=\textwidth]{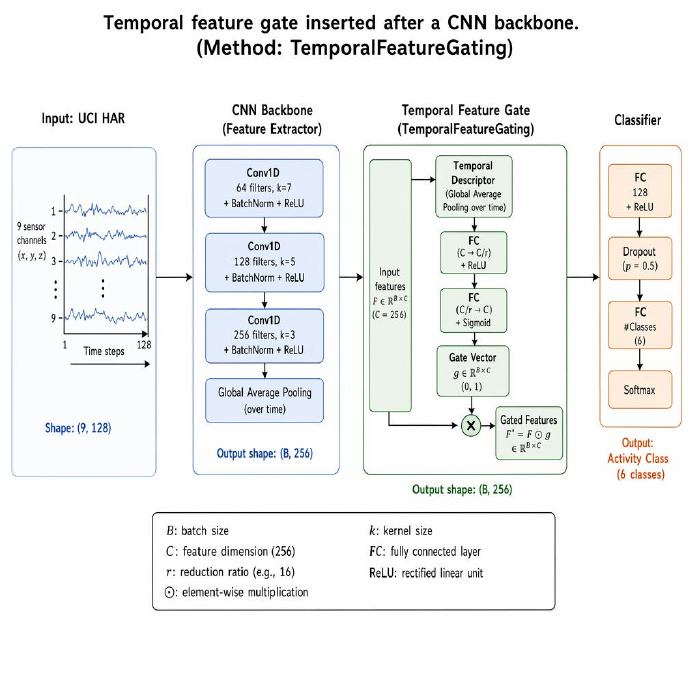}
  \caption{Architecture diagram for Profile B (Ablation-Focused Researcher).}
  \label{fig:case_study_profile_b}
\end{figure}
\paragraph{Stage I: Experiment blueprint.}
\begin{itemize}\setlength\itemsep{1pt}
  \item \textbf{title:} Temporal Feature Gating for Lightweight UCI HAR
  Classification.
  \item \textbf{proposed\_method.name:} TemporalFeatureGating.
  \item \textbf{proposed\_method.description:} A lightweight temporal gate
  is inserted after the 1D CNN feature extractor to predict importance
  weights over time steps before global pooling. The module is separated
  from the backbone so that its contribution can be isolated without
  changing the data loader, optimizer, or classifier.
  \item \textbf{proposed\_method.key\_components:} standard 1D CNN
  backbone; temporal gate with small bottleneck; weighted temporal
  pooling; compact classification head.
  \item \textbf{proposed\_method.architecture:} Input (9 channels $\times$
  128 time steps) $\rightarrow$ 1D CNN feature extractor $\rightarrow$
  temporal gate MLP $\rightarrow$ feature reweighting over time
  $\rightarrow$ global average pooling $\rightarrow$ linear classifier.
  \item \textbf{ablation\_groups:} gate removal ablation with
  \texttt{gate=None}; static-gate ablation with uniform temporal weights;
  bottleneck-size ablation with full versus tiny temporal gate.
\end{itemize}

\paragraph{Stage II: Coding output.}
The code is as follows.

\begin{lstlisting}[language=Python,
                   basicstyle=\ttfamily\small,
                   columns=fullflexible,
                   keepspaces=true,
                   showspaces=false,
                   showstringspaces=false,
                   breaklines=true]
def build_temporal_gate_encoder(
    num_channels: int,
    ablation: str = "full",
):
    """
    Compact encoder with a single reviewer-facing contribution module.
    The gate can be removed or simplified without changing the backbone.
    """
    gate = {
        "full":        TemporalGate(bottleneck=8, mode="learned"),
        "static_gate": TemporalGate(mode="uniform"),
        "tiny_gate":   TemporalGate(bottleneck=2, mode="learned"),
        "no_gate":     None,
    }[ablation]

    return TemporalGateEncoder(
        input_channels=num_channels,
        backbone="standard_1d_cnn",
        gate=gate,
    )
\end{lstlisting}

\paragraph{Stage III: Writing style.}
\begin{quote}
\itshape
The central claim is intentionally simple: a removable temporal gate
improves a 1D CNN only if its learned weighting survives direct removal,
static-gate, and tiny-gate ablations. Because the gate can be removed,
frozen to uniform weights, or compressed to a smaller bottleneck, the
same implementation directly tests whether learned temporal weighting
explains the observed accuracy--cost tradeoff.
\end{quote}

\subsection{Profile C: Benchmark-Driven Exploratory Researcher}
\label{app:case-study:profile-c}
\begin{figure}[!t]
  \centering
  \includegraphics[width=\textwidth]{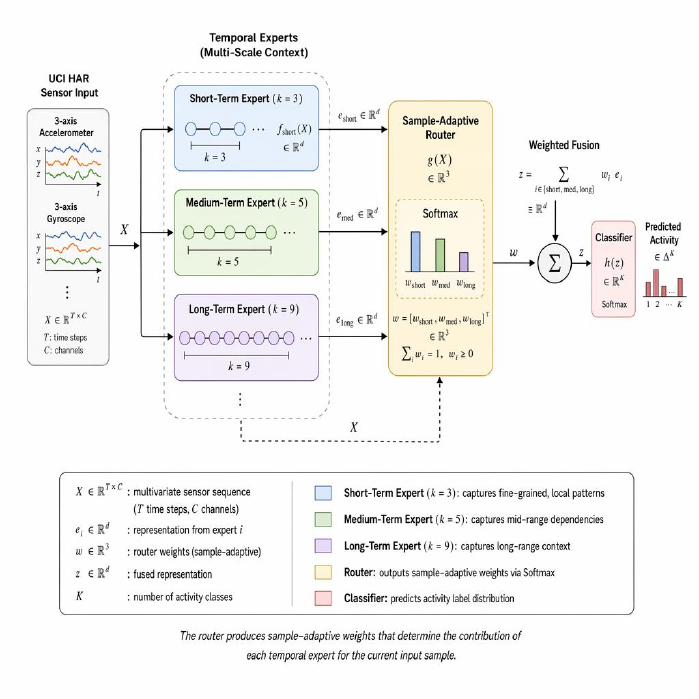}
  \caption{Architecture diagram for Profile C (Benchmark-Driven Exploratory Researcher).}
  \label{fig:case_study_profile_c}
\end{figure}
\paragraph{Stage I: Experiment blueprint.}
\begin{itemize}\setlength\itemsep{1pt}
  \item \textbf{title:} Temporal Evidence Routing for Benchmark-Oriented
  Sensor Classification.
  \item \textbf{proposed\_method.name:} TemporalEvidenceRouting.
  \item \textbf{proposed\_method.description:} A multi-expert temporal
  encoder uses short-, medium-, and long-range convolutional experts
  selected by a lightweight sample-adaptive router. The design turns
  multi-scale feature extraction into an input-dependent routing problem
  and supports broader benchmark-facing comparisons.
  \item \textbf{proposed\_method.key\_components:} short-range Conv1d
  expert; medium-range Conv1d expert; long-range Conv1d expert;
  sample-adaptive evidence router; shared classification head.
  \item \textbf{proposed\_method.architecture:} Input (9 channels $\times$
  128 time steps) $\rightarrow$ temporal experts with kernel sizes
  [3, 5, 9] $\rightarrow$ router predicts expert weights per sample
  $\rightarrow$ weighted expert fusion $\rightarrow$ global pooling
  $\rightarrow$ linear classifier.
  \item \textbf{ablation\_groups:} single-expert ablation; uniform-routing
  ablation; no-router feature-fusion ablation; routing-statistics
  analysis across activity classes.
\end{itemize}

\paragraph{Stage II: Coding output.}
The code is as follows.

\begin{lstlisting}[language=Python,
                   basicstyle=\ttfamily\small,
                   columns=fullflexible,
                   keepspaces=true,
                   showspaces=false,
                   showstringspaces=false,
                   breaklines=true]
def build_temporal_evidence_router(
    num_channels: int,
    routing: str = "sample_adaptive",
):
    """
    Benchmark-facing encoder with a named dynamic mechanism.
    The model routes each sample across temporal experts.
    """
    return TemporalEvidenceRouter(
        input_channels=num_channels,
        experts={
            "short_range": ConvExpert(kernel_size=3),
            "mid_range":   ConvExpert(kernel_size=5),
            "long_range":  ConvExpert(kernel_size=9),
        },
        router=EvidenceRouter(mode=routing),
        shared_classifier=True,
        export_routing_statistics=True,
    )
\end{lstlisting}

\paragraph{Stage III: Writing style.}
\begin{quote}
\itshape
We cast lightweight sensor classification as temporal evidence routing,
where each input dynamically selects the short-, medium-, or long-range
expert needed for strong benchmark performance. This design turns
multi-scale temporal modeling into a sample-adaptive decision, enabling a
stronger benchmark-facing comparison against fixed CNN, recurrent, and
Inception-style compact baselines.
\end{quote}

\section{Example of User Profile, Memory, and Skill}
\label{app:profile_example}

To make the abstract notions of \textit{User Profile}, \textit{Memory}, 
and \textit{Skill} concrete, we provide three representative examples 
produced by our system.

\begin{tcolorbox}[
  breakable,
  colback=blue!3,
  colframe=blue!50!black,
  boxrule=0.5pt,
  arc=2pt,
  left=6pt, right=6pt, top=6pt, bottom=6pt,
  title=\textbf{Profile A: Evidence-First Journal Scientist},
  fonttitle=\small\bfseries
]
\small
\textbf{User Profile.}
\begin{itemize}\setlength\itemsep{1pt}
  \item \textbf{Research preference:} exact reruns, explicit controls, reproducible ablations, and conservative methods built from standard PyTorch components.
  \item \textbf{Archetype:} ai4science\_journal.
  \item \textbf{Domain:} Time Series.
  \item \textbf{Method preference:} prefer exact reruns, explicit controls, and reproducible ablations over speculative novelty.
  \item \textbf{Risk preference:} low.
  \item \textbf{Baseline/ablation strictness:} very high.
  \item \textbf{Resource budget:} 1$\times$A100 80GB, 5 days.
  \item \textbf{Feasibility bias:} prefer explicit reproducibility steps, deterministic settings, and auditable experiment plans.
  \item \textbf{Writing tone:} highly restrained.
  \item \textbf{Claim strength:} conservative.
  \item \textbf{Section organization:} journal-style with dense evidence and careful limitations.
  \item \textbf{Venue style:} Nature/Springer journal.
  \item \textbf{LaTeX template preference:} nature\_springer.
  \item \textbf{Figure style:} composite scientific figure.
  \item \textbf{Caption style:} self-contained dense.
  \item \textbf{Priority feedback:} missing controls, weak reproducibility details, or hidden implementation variance.
  \item \textbf{Unacceptable errors:} overclaiming biological or physical conclusions, or under-specifying data provenance.
  \item \textbf{Router hints:} prefer profile over sparse memory; focus writing on conservative claims and dense evidence; focus planning on reproducibility and controlled compute.
  \item \textbf{Persona brief:} AI4Science persona that prioritizes reproducibility, exact reruns, and explicit experimental controls over flashy novelty.
\end{itemize}

\textbf{Profile summary.}\;
The user prioritizes exact reruns, deterministic evaluation, explicit controls, and auditable ablations. They prefer conservative methods built from standard PyTorch components over speculative architectural novelty. This profile tends to favor stable model designs with few dynamic decisions, and a restrained writing style that states narrow, evidence-grounded claims before broader interpretation.

\medskip
\textbf{Memory.}
\begin{itemize}\setlength\itemsep{2pt}
  \item \textbf{Memory 1.}
  \begin{itemize}\setlength\itemsep{1pt}
    \item \textbf{memory\_id:} mem-uci-har-baseline-protocol
    \item \textbf{memory\_type:} project\_context
    \item \textbf{source\_stage:} prior ideation and planning
    \item \textbf{topic\_scope:} lightweight time-series sensor classification on UCI HAR
    \item \textbf{content:} For UCI HAR, the first planning step should reproduce the compact baseline suite before evaluating a new method. The baseline suite should include a standard 1D CNN, a GRU, and InceptionTime-small. These baselines should use the same train/validation/test split, preprocessing pipeline, optimizer, learning rate, batch size, epoch budget, and evaluation script as the proposed method. Do not introduce a new model before establishing that the baseline implementations are executable and benchmark-comparable.
    \item \textbf{retrieval\_rationale:} Retrieved because the current task uses the same UCI HAR sensor-classification setup and the user profile strongly prioritizes exact reruns, controlled comparisons, and conservative evidence.
    \item \textbf{planning\_implication:} The blueprint should place baseline reproduction before proposed-method evaluation and should avoid changing the training protocol between baselines and the proposed model.
    \item \textbf{coding\_implication:} Implement one shared dataloader, one shared training loop, one shared evaluation function, and model-specific constructors only for the architecture difference.
    \item \textbf{writing\_implication:} Report the proposed method as a controlled extension after baseline reproduction, not as an unconstrained new architecture.
    \item \textbf{failure\_mode\_to\_avoid:} Do not compare the proposed method against baselines trained with different preprocessing, longer schedules, different random seeds, or missing validation controls.
  \end{itemize}

  \item \textbf{Memory 2.}
  \begin{itemize}\setlength\itemsep{1pt}
    \item \textbf{memory\_id:} mem-uci-har-fixed-controls
    \item \textbf{memory\_type:} decision\_history
    \item \textbf{source\_stage:} prior planning critique
    \item \textbf{topic\_scope:} UCI HAR controlled lightweight benchmark
    \item \textbf{content:} For this topic, experimental controls should remain fixed across all variants. The data split, sensor-channel preprocessing, normalization statistics, optimizer, learning rate, batch size, epoch budget, early-stopping rule, seed schedule, and metric computation should be identical for the 1D CNN, GRU, InceptionTime-small, proposed method, and all ablations. Any unavoidable implementation difference must be explicitly logged and justified.
    \item \textbf{retrieval\_rationale:} Retrieved because hidden implementation variance is especially harmful for this user's evidence-first profile and because UCI HAR is small enough that uncontrolled variance can dominate the reported gain.
    \item \textbf{planning\_implication:} The experiment blueprint should specify fixed controls directly rather than leaving them to the coding stage.
    \item \textbf{coding\_implication:} Put shared hyperparameters in a single config object and prevent per-model overrides unless the override is explicitly declared as an ablation.
    \item \textbf{analysis\_implication:} Report mean and standard deviation across repeated runs when possible, and separate accuracy changes from parameter-count or training-cost changes.
    \item \textbf{failure\_mode\_to\_avoid:} Do not let the proposed model receive extra epochs, different augmentation, different class weighting, or a tuned learning rate that the baselines do not receive.
  \end{itemize}

  \item \textbf{Memory 3.}
  \begin{itemize}\setlength\itemsep{1pt}
    \item \textbf{memory\_id:} mem-low-risk-temporal-module
    \item \textbf{memory\_type:} promising\_direction
    \item \textbf{source\_stage:} ideation reflection
    \item \textbf{topic\_scope:} compact wearable-sensor architectures
    \item \textbf{content:} A low-risk direction for UCI HAR is to add a small temporal module to a standard 1D CNN backbone instead of replacing the whole architecture. The module should test a narrow hypothesis about temporal feature extraction while preserving the baseline model's overall training behavior. Suitable directions include fixed multi-scale convolutions, shallow residual temporal blocks, or parameter-matched receptive-field changes. Avoid high-variance dynamic routing or attention-heavy mechanisms for this profile unless the task explicitly requires them.
    \item \textbf{retrieval\_rationale:} Retrieved because the user profile favors conservative standard PyTorch components and narrow claims over speculative architectural novelty.
    \item \textbf{planning\_implication:} Select a method that changes the temporal receptive-field structure while keeping the backbone recognizable as a 1D CNN.
    \item \textbf{coding\_implication:} Implement the new component as a small encoder block with a clear on/off or simplified variant for ablation.
    \item \textbf{writing\_implication:} Frame the contribution as a controlled hypothesis test rather than a broadly general architecture.
    \item \textbf{failure\_mode\_to\_avoid:} Do not propose a full replacement architecture whose gains cannot be attributed to one specific temporal modeling change.
  \end{itemize}
\end{itemize}

\textbf{Skill.}
\begin{itemize}\setlength\itemsep{2pt}
  \item \textbf{Skill 1.}
  \begin{itemize}\setlength\itemsep{1pt}
    \item \textbf{skill\_id:} skill-deterministic-experiment-contract
    \item \textbf{skill\_type:} planning\_and\_execution\_rule
    \item \textbf{name:} Write an explicit deterministic experiment contract.
    \item \textbf{when\_to\_apply:} Use when the user profile emphasizes reproducibility, exact reruns, auditability, or controlled scientific evidence.
    \item \textbf{procedure:} Specify fixed random seeds for Python, NumPy, PyTorch, and CUDA. Save the train/validation/test split indices to disk. Log package versions, CUDA version, device name, command-line arguments, config files, and git state when available. Enable deterministic PyTorch settings where practical and record any operations that remain nondeterministic. Make the same seed schedule available to baselines, proposed methods, and ablations.
    \item \textbf{planning\_effect:} The blueprint must include reproducibility controls as part of the experimental design, not as an optional implementation detail.
    \item \textbf{coding\_effect:} The generated code should expose a single \texttt{seed\_everything} utility, a saved-split loader, and a run manifest written with every experiment.
    \item \textbf{review\_check:} A reviewer should be able to rerun the same model variant and recover the same split, configuration, and evaluation protocol.
    \item \textbf{do\_not:} Do not claim reproducibility solely by saying that seeds are fixed; include saved splits, software logging, and shared evaluation scripts.
  \end{itemize}

  \item \textbf{Skill 2.}
  \begin{itemize}\setlength\itemsep{1pt}
    \item \textbf{skill\_id:} skill-one-factor-ablation-design
    \item \textbf{skill\_type:} experiment\_design\_rule
    \item \textbf{name:} Design one-factor-at-a-time ablations.
    \item \textbf{when\_to\_apply:} Use when the proposed method is a compact extension of a baseline and the main claim depends on attributing gains to a specific component.
    \item \textbf{procedure:} Start from the full proposed model. Define ablations that remove or simplify exactly one component at a time while keeping the data pipeline, optimizer, training schedule, model width where possible, and evaluation script fixed. Include a parameter-control variant when removing a component changes capacity substantially. Name each ablation by the changed factor rather than by a vague model nickname.
    \item \textbf{planning\_effect:} The blueprint should list ablation groups that map directly to the hypothesis being tested.
    \item \textbf{coding\_effect:} The implementation should use explicit variant flags rather than separate scripts that can silently diverge.
    \item \textbf{analysis\_effect:} Interpret improvements only when the full model beats the one-factor ablations under matched training conditions.
    \item \textbf{do\_not:} Do not combine multiple changes in one ablation, because this makes attribution impossible.
  \end{itemize}
\end{itemize}
\end{tcolorbox}

\begin{tcolorbox}[
  breakable,
  colback=orange!3,
  colframe=orange!60!black,
  boxrule=0.5pt,
  arc=2pt,
  left=6pt, right=6pt, top=6pt, bottom=6pt,
  title=\textbf{Profile B: Ablation-Focused Conference Researcher},
  fonttitle=\small\bfseries
]
\small
\textbf{User Profile.}
\begin{itemize}\setlength\itemsep{1pt}
  \item \textbf{Archetype:} nlp\_conference.
  \item \textbf{Domain:} Time Series.
  \item \textbf{Method preference:} pragmatic, compact methods with clean ablations, straightforward implementation paths, and reviewer-friendly framing.
  \item \textbf{Risk preference:} moderate.
  \item \textbf{Baseline/ablation strictness:} high.
  \item \textbf{Resource budget:} 1$\times$A100 80GB, 3 days.
  \item \textbf{Feasibility bias:} prefer methods feasible on small-to-medium compute budgets.
  \item \textbf{Writing tone:} restrained academic.
  \item \textbf{Claim strength:} moderate.
  \item \textbf{Section organization:} conference-style, direct and contribution-focused.
  \item \textbf{Venue style:} NeurIPS/ICLR conference.
  \item \textbf{LaTeX template preference:} conference\_template.
  \item \textbf{Figure style:} clean benchmark plots.
  \item \textbf{Caption style:} compact but informative.
  \item \textbf{Priority feedback:} needlessly complex methods or ablations that do not clarify the core contribution.
  \item \textbf{Unacceptable errors:} ignoring compute limits or skipping rigorous comparisons.
  \item \textbf{Router hints:} prefer profile over sparse memory; focus writing on direct contribution framing; focus planning on compact, ablatable, single-GPU methods.
  \item \textbf{Persona brief:} NLP conference persona that prefers pragmatic, ablatable methods with clean implementation paths and reviewer-friendly framing.
\end{itemize}

\textbf{Profile summary.}\;
The user prefers practical methods with clean ablations, straightforward implementation, and reviewer-friendly framing. They are open to moderate novelty when the contribution can be isolated as a compact module. This profile tends to favor code interfaces where the main component can be removed or simplified, and a concise writing style that foregrounds the core claim and its ablation evidence.

\medskip
\textbf{Memory.}
\begin{itemize}\setlength\itemsep{2pt}
  \item \textbf{Memory 1.}
  \begin{itemize}\setlength\itemsep{1pt}
    \item \textbf{memory\_id:} mem-plugin-module-reviewability
    \item \textbf{memory\_type:} promising\_direction
    \item \textbf{source\_stage:} prior ideation reflection
    \item \textbf{topic\_scope:} compact UCI HAR method design
    \item \textbf{content:} For a reviewer-friendly UCI HAR paper, a small plug-in module on top of a standard 1D CNN is easier to justify than a full architecture replacement. The contribution should be concentrated in one inspectable mechanism that can be named, removed, simplified, and compared against the same backbone. The method should remain easy to implement in PyTorch and feasible on a single A100 without large sweeps.
    \item \textbf{retrieval\_rationale:} Retrieved because the profile prefers practical conference-style contributions with clean ablations and straightforward implementation paths.
    \item \textbf{planning\_implication:} Select a method whose novelty is local and whose contribution can be explained in one paragraph.
    \item \textbf{coding\_implication:} Implement the contribution as a module argument or model variant flag rather than as a separate monolithic architecture.
    \item \textbf{writing\_implication:} Present the contribution as a compact module with a clear accuracy-cost tradeoff, not as a broad replacement for time-series modeling.
    \item \textbf{failure\_mode\_to\_avoid:} Do not create a method whose claimed novelty is spread across many small unrelated changes.
  \end{itemize}

  \item \textbf{Memory 2.}
  \begin{itemize}\setlength\itemsep{1pt}
    \item \textbf{memory\_id:} mem-direct-module-ablation-pressure
    \item \textbf{memory\_type:} decision\_history
    \item \textbf{source\_stage:} reviewer-style planning critique
    \item \textbf{topic\_scope:} ablation-centered method validation
    \item \textbf{content:} Reviewers will ask whether the proposed module itself causes the gain. The plan should therefore include direct module-removal, module-simplification, and static-replacement ablations. If the full method uses a learned temporal gate, include a no-gate variant, a uniform/static-gate variant, and a reduced-capacity gate variant. These ablations should share the same backbone, training loop, and evaluation code.
    \item \textbf{retrieval\_rationale:} Retrieved because the current profile values reviewer-facing evidence and because the selected topic is small enough to run direct ablations quickly.
    \item \textbf{planning\_implication:} The blueprint should make the ablation logic visible at the same time as the proposed method, not as an afterthought.
    \item \textbf{coding\_implication:} The model constructor should expose \texttt{full}, \texttt{no\_gate}, \texttt{static\_gate}, and \texttt{tiny\_gate} style variants through one unified interface.
    \item \textbf{analysis\_implication:} The final analysis should answer whether learned temporal weighting helps beyond added capacity or implementation noise.
    \item \textbf{failure\_mode\_to\_avoid:} Do not report only the full model and baselines; that leaves the central mechanism untested.
  \end{itemize}

  \item \textbf{Memory 3.}
  \begin{itemize}\setlength\itemsep{1pt}
    \item \textbf{memory\_id:} mem-uci-har-fast-benchmarkable-iteration
    \item \textbf{memory\_type:} project\_context
    \item \textbf{source\_stage:} benchmark planning memory
    \item \textbf{topic\_scope:} UCI HAR single-GPU experiments
    \item \textbf{content:} UCI HAR is small enough for fast iteration, so the experiment should prioritize a clean benchmarkable setup over large hyperparameter sweeps. Use a modest number of seeds or repeated runs if budget allows, but keep the main comparison simple. The plan should report accuracy, macro-F1, parameter count, runtime, and peak GPU memory so that reviewers can evaluate whether the compact module is worth its cost.
    \item \textbf{retrieval\_rationale:} Retrieved because the task requires feasibility on one GPU and the profile values practical, reviewer-friendly benchmarking.
    \item \textbf{planning\_implication:} Allocate compute to direct baselines and ablations before optional tuning.
    \item \textbf{coding\_implication:} Generate a unified training script with model variant flags and automatic resource logging.
    \item \textbf{writing\_implication:} Discuss both predictive performance and resource cost rather than only the best accuracy number.
    \item \textbf{failure\_mode\_to\_avoid:} Do not spend the compute budget on broad hyperparameter search that makes the main comparison harder to interpret.
  \end{itemize}
\end{itemize}

\textbf{Skill.}
\begin{itemize}\setlength\itemsep{2pt}
  \item \textbf{Skill 1.}
  \begin{itemize}\setlength\itemsep{1pt}
    \item \textbf{skill\_id:} skill-claim-to-ablation-map
    \item \textbf{skill\_type:} reviewer\_alignment\_rule
    \item \textbf{name:} Map every method claim to a direct ablation.
    \item \textbf{when\_to\_apply:} Use when the paper is intended to be reviewer-friendly and the contribution is a compact module or mechanism.
    \item \textbf{procedure:} List the method's claimed components. For each component, define a removal, replacement, or simplification ablation that tests whether that component is necessary. Keep the backbone, data pipeline, training schedule, and metrics fixed. Include the ablation name in the blueprint and ensure the coding interface can instantiate it directly.
    \item \textbf{planning\_effect:} The experiment plan should make the expected reviewer question and the corresponding ablation visible together.
    \item \textbf{coding\_effect:} Implement variants as stable flags in one constructor, such as \texttt{ablation="full"}, \texttt{ablation="no\_gate"}, \texttt{ablation="static\_gate"}, and \texttt{ablation="tiny\_gate"}.
    \item \textbf{writing\_effect:} Use the ablation results to support or narrow the central claim.
    \item \textbf{do\_not:} Do not introduce components that have no corresponding removal or replacement test.
  \end{itemize}

  \item \textbf{Skill 2.}
  \begin{itemize}\setlength\itemsep{1pt}
    \item \textbf{skill\_id:} skill-compact-module-framing
    \item \textbf{skill\_type:} writing\_and\_planning\_rule
    \item \textbf{name:} Frame the contribution as an inspectable plug-in module.
    \item \textbf{when\_to\_apply:} Use when the target style is a pragmatic conference paper and the method should be easy to benchmark.
    \item \textbf{procedure:} Name the module, state where it is inserted in the baseline, explain what signal it computes, and identify the direct comparison points. Avoid broad claims about replacing a whole modeling family. Emphasize that the same backbone can be evaluated with the module removed, frozen, or simplified.
    \item \textbf{planning\_effect:} The proposed method should be described around one compact mechanism rather than a bundle of unrelated improvements.
    \item \textbf{coding\_effect:} The module should be separable from the backbone and easy to instantiate, remove, or swap.
    \item \textbf{writing\_effect:} The introduction and method section should foreground what the module changes and why the ablations isolate that change.
    \item \textbf{do\_not:} Do not bury the core contribution inside implementation details that reviewers cannot isolate.
  \end{itemize}
\end{itemize}
\end{tcolorbox}

\begin{tcolorbox}[
  breakable,
  colback=green!3,
  colframe=green!50!black,
  boxrule=0.5pt,
  arc=2pt,
  left=6pt, right=6pt, top=6pt, bottom=6pt,
  title=\textbf{Profile C: Benchmark-Driven Exploratory Researcher},
  fonttitle=\small\bfseries
]
\small
\textbf{User Profile.}
\begin{itemize}\setlength\itemsep{1pt}
  \item \textbf{Archetype:} high\_novelty\_exploratory.
  \item \textbf{Domain:} Time Series.
  \item \textbf{Method preference:} strong benchmark coverage, broad comparisons, clear leaderboard-facing evidence, and a named mechanism with visible empirical upside.
  \item \textbf{Risk preference:} high, as long as novelty and compute constraints are explicit.
  \item \textbf{Baseline/ablation strictness:} medium-high.
  \item \textbf{Resource budget:} 2$\times$A100 80GB, 5 days.
  \item \textbf{Feasibility bias:} accept higher-risk proposals if novelty is clear and constraints are explicit.
  \item \textbf{Writing tone:} confident but disciplined.
  \item \textbf{Claim strength:} moderate-high.
  \item \textbf{Section organization:} concept-first with clear positioning.
  \item \textbf{Venue style:} benchmark-heavy conference.
  \item \textbf{LaTeX template preference:} conference\_template.
  \item \textbf{Figure style:} dense benchmark tables and comparison plots.
  \item \textbf{Caption style:} informative.
  \item \textbf{Priority feedback:} weak benchmark coverage or insufficient comparison breadth.
  \item \textbf{Unacceptable errors:} novelty claims unsupported by experiment or literature positioning.
  \item \textbf{Router hints:} prefer profile over sparse memory; focus writing on benchmark positioning; focus planning on comparison breadth and clearly named mechanisms.
  \item \textbf{Persona brief:} conference persona that values strong benchmark coverage, broad comparisons, and clear wins on established leaderboards.
\end{itemize}

\textbf{Profile summary.}\;
The user values broad benchmark coverage, strong comparisons, and clear leaderboard-facing evidence. They accept higher-risk proposals when novelty and compute constraints are explicit. This profile tends to favor more expressive mechanisms with named components or adaptive behavior, and a confident concept-first writing style anchored by benchmark comparisons.

\medskip
\textbf{Memory.}
\begin{itemize}\setlength\itemsep{2pt}
  \item \textbf{Memory 1.}
  \begin{itemize}\setlength\itemsep{1pt}
    \item \textbf{memory\_id:} mem-benchmark-breadth-sensor
    \item \textbf{memory\_type:} project\_context
    \item \textbf{source\_stage:} prior benchmark planning
    \item \textbf{topic\_scope:} benchmark-oriented lightweight sensor classification
    \item \textbf{content:} A method intended for strong benchmark positioning should not rely only on UCI HAR if another lightweight sensor dataset is feasible within budget. UCI HAR can remain the primary dataset, but the plan should reserve a secondary evaluation path for an additional wearable or sensor-classification dataset. If the secondary dataset cannot be completed, the paper should explicitly state this limitation and avoid broad leaderboard-style claims.
    \item \textbf{retrieval\_rationale:} Retrieved because the profile values broad comparisons and leaderboard-facing evidence rather than a single controlled case.
    \item \textbf{planning\_implication:} The blueprint should include a primary UCI HAR protocol and an optional secondary benchmark protocol with the same model variants.
    \item \textbf{coding\_implication:} Design the dataloader and training script so that dataset selection is configurable rather than hard-coded to UCI HAR.
    \item \textbf{analysis\_implication:} Separate single-dataset findings from cross-dataset findings.
    \item \textbf{failure\_mode\_to\_avoid:} Do not claim benchmark generality from one dataset unless comparison breadth is actually present.
  \end{itemize}

  \item \textbf{Memory 2.}
  \begin{itemize}\setlength\itemsep{1pt}
    \item \textbf{memory\_id:} mem-named-concept-positioning
    \item \textbf{memory\_type:} writing\_context
    \item \textbf{source\_stage:} prior paper review
    \item \textbf{topic\_scope:} concept-first conference writing
    \item \textbf{content:} The introduction should make the core concept easy to remember and distinguish from standard CNN, recurrent, and Inception-style baselines. A benchmark-facing method benefits from a named mechanism that captures the central modeling idea. The name should correspond to an actual architectural behavior, not just a rebranding of ordinary convolution or pooling.
    \item \textbf{retrieval\_rationale:} Retrieved because the user profile prefers confident concept-first writing and visible empirical upside.
    \item \textbf{planning\_implication:} Select or name a method around a mechanism that can be explained as more than a local implementation tweak.
    \item \textbf{coding\_implication:} Expose internal quantities, such as routing weights or expert usage, when they support the named mechanism.
    \item \textbf{writing\_implication:} Open the paper with the concept and then connect it to benchmark evidence, resource cost, and ablations.
    \item \textbf{failure\_mode\_to\_avoid:} Do not use a catchy method name if the architecture does not provide a distinct mechanism behind the name.
  \end{itemize}

  \item \textbf{Memory 3.}
  \begin{itemize}\setlength\itemsep{1pt}
    \item \textbf{memory\_id:} mem-benchmark-claims-need-comparison-breadth
    \item \textbf{memory\_type:} decision\_history
    \item \textbf{source\_stage:} benchmark result reflection
    \item \textbf{topic\_scope:} conference-style empirical claims
    \item \textbf{content:} Benchmark-heavy claims require comparison breadth, not only a single best-number improvement. The evaluation should include standard compact baselines, the strongest relevant prior-round compact variant when available, resource metrics, and failure or sensitivity analysis. Strong claims should be conditional on both performance and cost. If the method is adaptive, report whether the adaptive behavior is meaningful rather than only reporting final accuracy.
    \item \textbf{retrieval\_rationale:} Retrieved because the current profile accepts higher-risk methods only when novelty and benchmark evidence are explicit.
    \item \textbf{planning\_implication:} Add comparisons beyond the minimum baseline suite when compute permits, and include resource reporting as part of the main table.
    \item \textbf{coding\_implication:} Log routing statistics, parameter counts, runtime, and memory usage in addition to predictive metrics.
    \item \textbf{writing\_implication:} Use confident language only when benchmark breadth and diagnostic evidence support it.
    \item \textbf{failure\_mode\_to\_avoid:} Do not turn one positive result into a broad claim about general sensor-classification superiority.
  \end{itemize}
\end{itemize}

\textbf{Skill.}
\begin{itemize}\setlength\itemsep{2pt}
  \item \textbf{Skill 1.}
  \begin{itemize}\setlength\itemsep{1pt}
    \item \textbf{skill\_id:} skill-concept-level-mechanism
    \item \textbf{skill\_type:} ideation\_and\_writing\_rule
    \item \textbf{name:} Convert a local architectural change into a real concept-level mechanism.
    \item \textbf{when\_to\_apply:} Use when the profile favors novelty, benchmark positioning, and a memorable contribution.
    \item \textbf{procedure:} Identify the actual behavior introduced by the method, name that behavior, and ensure the architecture exposes evidence that the behavior occurs. The name should describe a mechanism such as routing, evidence selection, scale specialization, or adaptive aggregation. The method section should connect the name to equations or code-level components.
    \item \textbf{planning\_effect:} The selected idea should have a distinct mechanism that can support a concept-first paper narrative.
    \item \textbf{coding\_effect:} Implement diagnostic outputs that make the mechanism observable, such as routing weights, expert usage, or scale-selection entropy.
    \item \textbf{writing\_effect:} The introduction can use a stronger hook because the named concept corresponds to measurable behavior.
    \item \textbf{do\_not:} Do not rename a standard block without changing or measuring its behavior.
  \end{itemize}

  \item \textbf{Skill 2.}
  \begin{itemize}\setlength\itemsep{1pt}
    \item \textbf{skill\_id:} skill-benchmark-claim-evidence-package
    \item \textbf{skill\_type:} evaluation\_and\_writing\_rule
    \item \textbf{name:} Pair strong claims with benchmark, resource, and failure evidence.
    \item \textbf{when\_to\_apply:} Use when the paper makes benchmark-facing or novelty-forward claims.
    \item \textbf{procedure:} For every strong claim, provide a table or diagnostic result that supports it. Pair performance tables with parameter count, runtime, memory, and ablation results. Include at least one failure case, sensitivity analysis, or limitation when the method is more complex than a baseline. If the method uses adaptive behavior, report diagnostics showing when and how adaptation occurs.
    \item \textbf{planning\_effect:} The blueprint should include comparison breadth, resource reporting, and diagnostic analysis as first-class experiment outputs.
    \item \textbf{coding\_effect:} The generated scripts should automatically save metrics, resource logs, and mechanism-specific diagnostics.
    \item \textbf{writing\_effect:} Claims should be confident but conditional on the observed benchmark and diagnostic evidence.
    \item \textbf{do\_not:} Do not write leaderboard-style conclusions without resource and failure analysis.
  \end{itemize}
\end{itemize}
\end{tcolorbox}

\section{Research Topics Specification}
\label{app:topics}

This section lists the \textbf{20 research topics} used in our
evaluation. The topics span seven domains and cover a diverse set of
task types, modalities, and evaluation setups. The complete list is
given in Listing~\ref{lst:topics-json}.

\lstset{language=json}
\begin{lstlisting}[
    caption={Full JSON specification of the 20 research topics used in our evaluation.},
    label={lst:topics-json},
    captionpos=b,
    numbers=none,
    breaklines=true,
    breakatwhitespace=true,
    columns=fullflexible,
    keepspaces=true,
    basicstyle=\ttfamily\footnotesize,
    frame=single,
    xleftmargin=0.5em,
    framexleftmargin=0.5em
]
[
  {
    "question_id": "nlp_biomed_qa",
    "domain": "NLP",
    "difficulty": "incremental_innovation",
    "background": "Biomedical QA systems already perform reasonably well on PubMedQA, but lightweight improvements with clean ablations and reproducible training are still valuable.",
    "problem_statement": "Design a practical method for improving PubMedQA under limited compute while keeping the implementation easy to reproduce.",
    "baselines": ["BioBERT", "PubMedBERT", "instruction-tuned biomedical QA baseline"],
    "datasets": ["PubMedQA"],
    "user_requirements": "Generate a new idea and an implementation-oriented plan. Keep the method lightweight, ablatable, and reproducible.",
    "extra_context": "Prefer methods that fit within a modest single-node budget and can be compared fairly against standard biomedical QA baselines."
  },
  {
    "question_id": "nlp_short_text_cls",
    "domain": "NLP",
    "difficulty": "incremental_innovation",
    "background": "Short-text classification benchmarks are mature, but small improvements that reduce compute and keep the stack simple remain useful for reproducible evaluation.",
    "problem_statement": "Propose a lightweight method for improving short-text classification quality without introducing a heavy training pipeline.",
    "baselines": ["DistilBERT", "BERT-base", "linear bag-of-words classifier"],
    "datasets": ["AG News", "SST-2"],
    "user_requirements": "Produce a practical idea and an executable plan. Favor compact architectures, clean ablations, and fast iteration.",
    "extra_context": "The project should be feasible on a single GPU with small batch sizes and should avoid retrieval-heavy or multi-stage systems."
  },
  {
    "question_id": "nlp_sentence_pair_cls",
    "domain": "NLP",
    "difficulty": "incremental_innovation",
    "background": "Sentence-pair benchmarks are easy to fine-tune and compare, making them a good testbed for compact modeling ideas rather than large-scale engineering.",
    "problem_statement": "Design a lightweight method for improving sentence-pair classification or matching quality without adding a heavy retrieval or multi-stage stack.",
    "baselines": ["DistilBERT", "BERT-base", "Siamese bi-encoder baseline"],
    "datasets": ["MRPC", "RTE"],
    "user_requirements": "Return a practical research idea and an executable implementation plan. Favor compact modules, fair baselines, and short training cycles.",
    "extra_context": "The method should remain small enough for a single-GPU run and should allow clear ablations over standard sentence-pair baselines."
  },
  {
    "question_id": "cv_small_image_cls",
    "domain": "CV",
    "difficulty": "incremental_innovation",
    "background": "Small-image classification tasks are easy to run and compare, making them suitable for testing whether the agent can propose reproducible improvements rather than large-scale engineering tricks.",
    "problem_statement": "Design a lightweight image-classification method that improves small-image benchmarks without relying on oversized backbones or expensive pretraining.",
    "baselines": ["ResNet-18", "MobileNetV3-small", "ViT-tiny"],
    "datasets": ["CIFAR-10", "FashionMNIST"],
    "user_requirements": "Return a novel but practical method and a benchmarkable implementation plan. Keep the method compact, ablatable, and easy to train.",
    "extra_context": "Prefer methods that can finish a meaningful run quickly on a single GPU and use standard torchvision-style tooling."
  },
  {
    "question_id": "multimodal_efficiency",
    "domain": "Multimodal",
    "difficulty": "nontrivial_recomposition",
    "background": "Compact multimodal systems often trade off quality against latency and systems complexity, especially when evaluated under strict deployment budgets.",
    "problem_statement": "Propose a systems-aware multimodal method that improves benchmark quality without introducing an impractical training or serving stack.",
    "baselines": ["compact VLM baseline", "late-fusion multimodal baseline"],
    "datasets": ["MMMU", "ScienceQA"],
    "user_requirements": "Produce a novel idea, an executable plan, and a benchmarkable implementation path. Avoid overly fragile or heavy multi-stage designs.",
    "extra_context": "Reward methods with clear component interfaces, fair comparisons, and realistic implementation scope. Keep the stack compact enough for a single-GPU experiment."
  },
  {
    "question_id": "tabular_budgeted_cls",
    "domain": "Tabular ML",
    "difficulty": "incremental_innovation",
    "background": "Tabular classification remains a strong testbed for low-cost experimentation because datasets are small, baselines are well understood, and implementation cycles are fast.",
    "problem_statement": "Design a lightweight tabular-learning method that improves standard tabular baselines without relying on large ensembles or expensive feature engineering.",
    "baselines": ["XGBoost", "TabTransformer", "MLP baseline"],
    "datasets": ["Adult", "CoverType"],
    "user_requirements": "Generate an idea and an implementation-oriented plan that is simple, fast, and easy to compare against standard baselines.",
    "extra_context": "Favor methods that can be trained in a short wall-clock time, with clear ablations and no dependence on external retrieval or long preprocessing pipelines."
  },
  {
    "question_id": "tabular_regression",
    "domain": "Tabular ML",
    "difficulty": "incremental_innovation",
    "background": "Small tabular regression problems are cheap to run and easy to diagnose, so they are useful for testing whether the system can make focused improvements under tight resource budgets.",
    "problem_statement": "Propose a lightweight tabular-regression method that improves standard regression baselines without resorting to large ensembles or expensive feature engineering.",
    "baselines": ["XGBoost regressor", "CatBoost regressor", "MLP regressor baseline"],
    "datasets": ["California Housing", "Energy Efficiency"],
    "user_requirements": "Return a practical idea and a runnable implementation plan. Favor compact models, clear ablations, and fast turnaround.",
    "extra_context": "The project should be easy to implement with sklearn-style preprocessing or a small PyTorch model and should finish quickly on a single GPU or CPU-backed node."
  },
  {
    "question_id": "timeseries_sensor_cls",
    "domain": "Time Series",
    "difficulty": "incremental_innovation",
    "background": "Human activity recognition and other compact time-series tasks are lightweight enough for repeated experimentation while still requiring nontrivial modeling choices.",
    "problem_statement": "Propose a lightweight time-series classification method that improves compact sensor benchmarks without using a large or highly specialized model stack.",
    "baselines": ["1D CNN baseline", "GRU baseline", "InceptionTime-small"],
    "datasets": ["UCI HAR"],
    "user_requirements": "Return a practical method and a reproducible implementation plan. Keep the approach compact, interpretable, and easy to benchmark.",
    "extra_context": "Prefer methods that can be implemented with standard PyTorch components and evaluated quickly on a single GPU."
  },
  {
    "question_id": "graph_node_cls",
    "domain": "Graph ML",
    "difficulty": "incremental_innovation",
    "background": "Node classification on citation graphs is a lightweight setting for probing whether the system can propose meaningful graph-model refinements without large-scale infrastructure.",
    "problem_statement": "Design a lightweight graph-learning method that improves standard node-classification baselines while keeping the implementation simple and reproducible.",
    "baselines": ["GCN", "GraphSAGE", "GAT"],
    "datasets": ["Cora", "Citeseer"],
    "user_requirements": "Generate a new idea and an executable plan. Prefer simple message-passing modifications, clean ablations, and modest compute cost.",
    "extra_context": "Avoid large graph pretraining or multi-stage pipelines. The project should be runnable quickly with a standard single-GPU setup."
  },
  {
    "question_id": "audio_keyword_cls",
    "domain": "Audio",
    "difficulty": "incremental_innovation",
    "background": "Keyword spotting on short audio clips is a compact benchmark family that is cheap to train, easy to compare, and useful for testing whether the system can make practical efficiency-oriented improvements.",
    "problem_statement": "Design a lightweight audio-classification method that improves keyword spotting quality without depending on a large speech model or a complex preprocessing pipeline.",
    "baselines": ["CNN keyword-spotting baseline", "CRNN baseline", "tiny conformer baseline"],
    "datasets": ["SpeechCommands"],
    "user_requirements": "Return a practical idea and an executable plan. Keep the model compact, the preprocessing standard, and the training budget modest.",
    "extra_context": "Favor torchaudio-compatible pipelines and methods that can finish a meaningful comparison quickly on a single GPU."
  },
  {
    "question_id": "nlp_token_cls",
    "domain": "NLP",
    "difficulty": "incremental_innovation",
    "background": "Token-level sequence labeling tasks remain one of the cleanest settings for testing compact contextual modeling ideas and fair ablations under limited compute.",
    "problem_statement": "Design a lightweight token-classification method that improves standard named-entity recognition benchmarks without adding a large pipeline or external retrieval system.",
    "baselines": ["DistilBERT token classifier", "BERT-base token classifier", "BiLSTM-CRF baseline"],
    "datasets": ["CoNLL-2003", "WNUT17"],
    "user_requirements": "Return a practical idea and an executable implementation plan. Favor compact sequence modules, clean ablations, and short fine-tuning runs.",
    "extra_context": "The full project should run on a single GPU and should stay close to standard Hugging Face token-classification tooling."
  },
  {
    "question_id": "nlp_extractive_qa",
    "domain": "NLP",
    "difficulty": "incremental_innovation",
    "background": "Extractive QA is mature enough that strong baselines are available, but lightweight methods that improve calibration or answer localization remain useful and easy to benchmark.",
    "problem_statement": "Propose a lightweight extractive QA method that improves standard span-selection baselines without relying on retrieval-heavy or multi-stage systems.",
    "baselines": ["DistilBERT QA baseline", "BERT-base QA baseline", "RoBERTa-base QA baseline"],
    "datasets": ["SQuAD v1.1", "NewsQA"],
    "user_requirements": "Produce a practical method and an implementation-oriented plan. Keep the method compact, reproducible, and easy to compare against standard QA baselines.",
    "extra_context": "Prefer single-model approaches that can be trained on one GPU and evaluated with standard extractive QA metrics."
  },
  {
    "question_id": "cv_finegrained_cls",
    "domain": "CV",
    "difficulty": "incremental_innovation",
    "background": "Fine-grained visual classification is harder than small-image classification but still manageable on compact datasets, making it a good benchmark for lightweight representation improvements.",
    "problem_statement": "Design a lightweight fine-grained image-classification method that improves compact benchmarks without requiring a large pretrained vision backbone.",
    "baselines": ["ResNet-18", "EfficientNet-B0", "ViT-tiny"],
    "datasets": ["Oxford-IIIT Pets", "Flowers102"],
    "user_requirements": "Return a practical method and a benchmarkable implementation plan. Favor compact modules, fair baselines, and simple training code.",
    "extra_context": "The benchmark should remain feasible on a single GPU with standard torchvision or timm tooling and should support clear ablations."
  },
  {
    "question_id": "cv_multilabel_cls",
    "domain": "CV",
    "difficulty": "incremental_innovation",
    "background": "Compact multi-label image classification is a useful stress test for calibration and feature-sharing ideas while staying much cheaper than large-scale detection pipelines.",
    "problem_statement": "Propose a lightweight multi-label image-classification method that improves standard baselines without introducing a heavy detection or segmentation stack.",
    "baselines": ["ResNet-18 multi-label baseline", "MobileNetV3 multi-label baseline", "ViT-tiny multi-label baseline"],
    "datasets": ["Pascal VOC 2007"],
    "user_requirements": "Generate a practical idea and an executable implementation plan. Keep the model small, the training pipeline standard, and the ablations clean.",
    "extra_context": "Prefer methods that can run with standard image-classification backbones and sigmoid multi-label heads on a single GPU."
  },
  {
    "question_id": "tabular_imbalance_cls",
    "domain": "Tabular ML",
    "difficulty": "incremental_innovation",
    "background": "Imbalanced tabular classification is common in real applications and is still easy to benchmark with compact models and short training loops.",
    "problem_statement": "Design a lightweight tabular method for improving imbalanced classification without relying on large ensembles, costly resampling pipelines, or heavy AutoML stacks.",
    "baselines": ["XGBoost", "LightGBM", "MLP baseline"],
    "datasets": ["Credit Card Fraud", "Telco Churn"],
    "user_requirements": "Return a practical method and a runnable plan. Favor compact architectures, fair imbalance-aware metrics, and fast experimentation.",
    "extra_context": "The project should be implementable with sklearn-style preprocessing and short training jobs on CPU or a single modest GPU."
  },
  {
    "question_id": "tabular_missing_value_cls",
    "domain": "Tabular ML",
    "difficulty": "incremental_innovation",
    "background": "Missing values are a realistic source of difficulty in tabular learning and provide a clean testbed for lightweight robustness ideas under short iteration cycles.",
    "problem_statement": "Propose a lightweight tabular-classification method that improves robustness to missing-value patterns without relying on expensive imputation ensembles or large stacked models.",
    "baselines": ["XGBoost", "TabTransformer", "MLP baseline with imputation"],
    "datasets": ["Adult", "Higgs Small"],
    "user_requirements": "Produce a practical idea and an executable implementation plan. Keep preprocessing simple, ablations clear, and compute modest.",
    "extra_context": "Prefer methods that can be implemented with simple masking or feature-gating ideas and benchmarked quickly on standard tabular datasets."
  },
  {
    "question_id": "timeseries_ecg_cls",
    "domain": "Time Series",
    "difficulty": "incremental_innovation",
    "background": "Compact ECG and UCR-style sequence benchmarks are cheap to run and make it easy to compare lightweight temporal architectures without specialized infrastructure.",
    "problem_statement": "Design a lightweight time-series classification method that improves compact ECG-style benchmarks without using a large transformer stack or custom hardware assumptions.",
    "baselines": ["1D CNN baseline", "GRU baseline", "InceptionTime-small"],
    "datasets": ["ECG200", "FordA"],
    "user_requirements": "Return a practical method and a reproducible implementation plan. Favor small temporal modules, interpretable ablations, and fast turnaround.",
    "extra_context": "The full experiment should remain compatible with plain PyTorch and run comfortably on a single GPU."
  },
  {
    "question_id": "graph_link_pred",
    "domain": "Graph ML",
    "difficulty": "incremental_innovation",
    "background": "Lightweight link prediction on citation or collaboration graphs is a compact graph-learning setting that still allows meaningful architectural comparison and ablation.",
    "problem_statement": "Propose a lightweight graph-learning method that improves standard link-prediction baselines while keeping the codebase simple and easy to reproduce.",
    "baselines": ["GCN encoder + dot-product decoder", "GraphSAGE encoder + MLP decoder", "GAT encoder baseline"],
    "datasets": ["Cora", "Citeseer"],
    "user_requirements": "Generate a new idea and an executable plan. Favor simple neighborhood or edge-scoring modifications, clean ablations, and modest compute cost.",
    "extra_context": "Avoid large graph pretraining and keep the project runnable on a single GPU with standard PyTorch Geometric tooling."
  },
  {
    "question_id": "audio_emotion_cls",
    "domain": "Audio",
    "difficulty": "incremental_innovation",
    "background": "Small audio emotion benchmarks are compact enough for repeated experimentation and are a useful testbed for lightweight temporal and spectral feature-learning ideas.",
    "problem_statement": "Design a lightweight audio-classification method that improves small emotion-recognition benchmarks without depending on a large speech foundation model.",
    "baselines": ["CNN spectrogram baseline", "CRNN baseline", "tiny conformer baseline"],
    "datasets": ["RAVDESS", "CREMA-D"],
    "user_requirements": "Return a practical idea and an executable plan. Keep preprocessing standard, the model compact, and the implementation reproducible.",
    "extra_context": "Favor torchaudio-compatible pipelines, short training cycles, and methods that can be compared fairly on a single GPU."
  },
  {
    "question_id": "multimodal_hateful_memes",
    "domain": "Multimodal",
    "difficulty": "nontrivial_recomposition",
    "background": "Compact multimodal classification tasks are useful for testing whether the system can improve cross-modal fusion quality without leaning on a very large vision-language model.",
    "problem_statement": "Propose a lightweight multimodal classification method that improves compact vision-language benchmarks without introducing a heavy multi-stage or large-model serving stack.",
    "baselines": ["late-fusion multimodal baseline", "CLIP linear-probe baseline", "compact VLM baseline"],
    "datasets": ["Hateful Memes"],
    "user_requirements": "Produce a novel idea, an executable plan, and a benchmarkable implementation path. Keep the model compact, the fusion strategy interpretable, and the evaluation fair.",
    "extra_context": "Reward methods with clean image-text interfaces and realistic single-GPU training requirements."
  }
]
\end{lstlisting}



\end{document}